\documentclass{article}
\usepackage{icip,amsmath,graphicx}
\usepackage{adjustbox}
\usepackage{booktabs}
\usepackage{multirow}
\usepackage{bibunits}
\usepackage{eso-pic}

\newcommand{\IEEEcopyrightnotice}{%
\AddToShipoutPictureFG*{%
  \AtPageLowerLeft{%
    \raisebox{0.22in}{%
      \makebox[\paperwidth][c]{%
        \parbox{0.94\paperwidth}{%
          \tiny
          Copyright 2026 IEEE. Published in 2026 IEEE International Conference on Image Processing (ICIP), scheduled for 13--17 September 2026 in Tampere, Finland. Personal use of this material is permitted. Permission from IEEE must be obtained for all other uses, in any current or future media, including reprinting/republishing this material for advertising or promotional purposes, creating new collective works, for resale or redistribution to servers or lists, or reuse of any copyrighted component of this work in other works.
        }%
      }%
    }%
  }%
}%
}

\title{Clip-level Uncertainty and Temporal-aware Active Learning for End-to-End Multi-Object Tracking}

\name{\shortstack{
Riku Inoue\qquad Shogo Sato\qquad Kazuhiko Murasaki\qquad Tomoyasu Shimada\\
Toshihiko Nishimura\qquad Ryuichi Tanida
}}

\address{NTT, Inc.,  Kanagawa,  Japan}

\begin{document}

\maketitle
\IEEEcopyrightnotice

\begin{bibunit}[icip]

\begin{abstract}

\noindent Multi-Object Tracking (MOT) in dynamic environments relies on robust temporal reasoning to maintain consistent object identities over time.
Transformer-based end-to-end MOT models achieve strong performance by explicitly modeling temporal dependencies, yet training them requires extensive bounding-box and identity annotations.
Given the high labeling cost and strong redundancy in videos, Active Learning (AL) is an effective approach to improve annotation efficiency. 
However, existing AL methods for MOT primarily operate at the frame level, which is structurally misaligned with modern end-to-end trackers whose inference and training rely on multi-frame clips.
To bridge this gap, we formulate clip-level active learning and propose Clip-level Uncertainty and Temporal-aware Active Learning (CUTAL). 
In contrast to frame-based approaches, CUTAL scores each clip using uncertainty metrics derived from multi-frame predictions to capture inter-frame correspondence ambiguities, while enforcing temporal diversity to select an informative and non-redundant subset. 
Experiments show that CUTAL achieves stronger overall performance than baselines at the same label budgets across MeMOTR and SambaMOTR.
Notably, CUTAL achieves performance comparable to full supervision for MeMOTR on both datasets using only 50\% of the labeled training data.

\end{abstract}


\section{Introduction}
\label{sec:intro}

\noindent Multi-Object Tracking (MOT) is fundamental to visual perception in dynamic environments, requiring simultaneous detection and consistent identity maintenance over time.
It is widely used in applications such as autonomous driving~\cite{yu2020bdd100k} and sports analytics~\cite{cui2023sportsmot}.
Recent MOT benchmarks feature crowded scenes with visually similar targets, frequent occlusions, and highly nonlinear motion~\cite{sun2022dancetrack,cui2023sportsmot}, where robust tracking requires explicit temporal reasoning beyond frame-wise heuristics.

To address these challenges, end-to-end MOT methods have been proposed, which propagate track queries across frames and perform joint temporal reasoning over video sequences~\cite{meinhardt2022trackformer,zeng2022motr,gao2023memotr,segu2025samba}.
However, training these models is annotation-intensive because MOT requires frame-wise boxes with temporally consistent identity labels.
Furthermore, while videos contain strong temporal redundancy, existing pipelines rarely prioritize which temporal segments to label.
Consequently, limited budgets are often wasted on redundant clips rather than on the most informative ones.

\begin{figure}[t]
    \centering
    \includegraphics[width=0.87\linewidth]{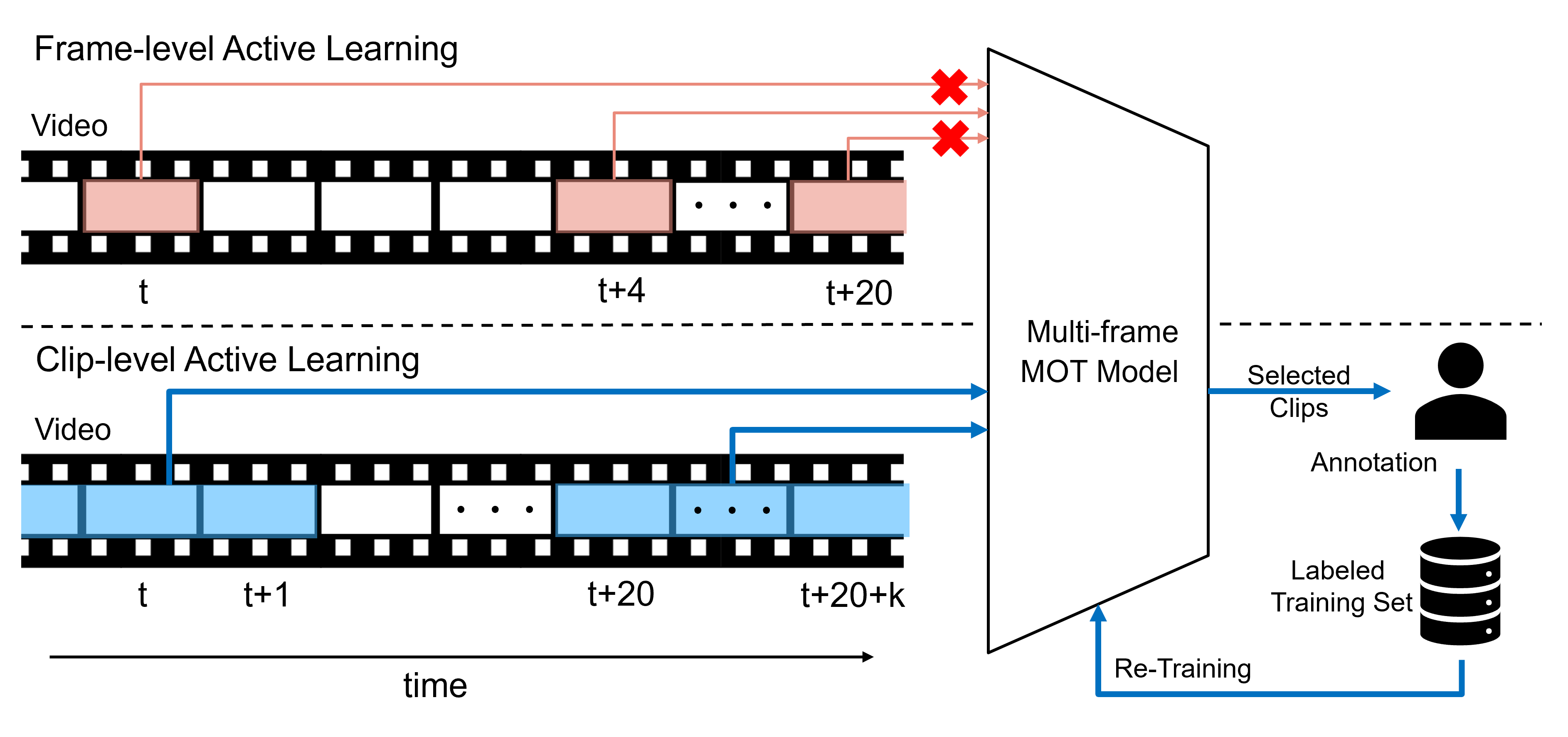}
    \caption{\textbf{Frame-level vs.\ clip-level active learning.}
    This conceptual figure contrasts isolated frame acquisition with temporally ordered clip acquisition.
    Clip-level acquisition preserves the multi-frame context used by end-to-end MOT training.}
    \label{fig:al_framework}
\end{figure}

Active Learning (AL) offers a principled way to improve annotation efficiency by selecting informative samples under a limited budget~\cite{wang2016cost,sener2017active,ash2019deep}.
Prior work on active learning for video tasks~\cite{rana2022all,rana2023hybrid}, including MOT~\cite{li2023heterogeneous}, has typically formulated acquisition at the frame level.
Despite their effectiveness, such frame-centric formulations are fundamentally misaligned with multi-frame trackers that learn temporal correspondence from multi-frame clips (Figure~\ref{fig:al_framework}).

To bridge this gap, we formulate clip-level active learning and propose Clip-level Uncertainty and Temporal-aware Active Learning (CUTAL).
We define the acquisition unit as a fixed-length clip, preserving a coherent multi-frame temporal context for multi-frame end-to-end trackers.
CUTAL assigns each unlabeled clip an uncertainty score computed from sequential multi-frame predictions, capturing ambiguous inter-frame correspondence and unstable identity assignments.
It then selects a non-redundant batch under a fixed annotation budget via temporal diversity sampling that favors clips far from already labeled ones along the timeline.

Our contributions are summarized as follows:
\begin{itemize}
\item We formulate clip-level active learning to resolve the structural mismatch between frame-level selection and the clip-based temporal reasoning of end-to-end trackers. 
\item We propose CUTAL, which evaluates clips using uncertainty metrics tailored to capture inter-frame correspondence ambiguities, integrated with diversity sampling to maximize information gain.
\item Extensive experiments on DanceTrack and SportsMOT demonstrate that CUTAL outperforms baselines. 
Notably, it enables MeMOTR to achieve comparable performance to full supervision using only 50\% of the labeled data.
\end{itemize}

\section{Related Work}
\label{sec:related_work}

\noindent\textbf{Multi-Object Tracking.}
\noindent Early MOT research was dominated by the tracking-by-detection paradigm~\cite{zhang2022bytetrack}, where a strong detector~\cite{ge2021yolox} produces per-frame detections and data association is handled as a post-processing step using IoU-based matching~\cite{bewley2016simple} and re-identification features~\cite{zhang2021fairmot}.
While effective on pedestrian-centric benchmarks such as MOT17~\cite{milan2016mot16}, frame-wise detection and heuristic association become unreliable in crowded scenes with similar appearances, frequent occlusions, and complex motion~\cite{sun2022dancetrack,cui2023sportsmot}.
To explicitly model temporal dependencies, end-to-end MOT methods propagate track queries across frames and jointly optimize detection and tracking within a single framework~\cite{meinhardt2022trackformer,zeng2022motr}.
MeMOTR~\cite{gao2023memotr} extends this line with per-track long-term memory and a temporal interaction module to inject longer temporal cues into track embeddings.
More recently, SambaMOTR~\cite{segu2025samba} replaces query propagation module with Samba, a synchronized set-of-sequences model built on selective state-space models~\cite{gu2024mamba}, and employs MaskObs to prevent unreliable observations from corrupting propagated queries.

\noindent\textbf{Active Learning.}
\noindent AL has been widely studied through uncertainty~\cite{wang2016cost}, diversity~\cite{sener2017active}, and hybrid strategies~\cite{ash2019deep}.
For video tasks, prior work often performs frame-level acquisition to reduce redundancy, selecting a sparse set of informative frames within each video~\cite{rana2022all,rana2023hybrid}.
In MOT-specific settings, HD-AMOT~\cite{li2023heterogeneous} casts informative frame selection as a Markov decision process and learns a batch sampling policy from heterogeneous geometric and semantic cues.
SPAM~\cite{cetintas2024spamming} develops a video label engine that chains synthetic pre-training and pseudo-label self-training, and actively requests human input for uncertain detection filtering or association decisions within a hierarchical graph model.
In contrast to these formulations, we target multi-frame end-to-end trackers and perform acquisition at the clip level to match their inference and training structure, using uncertainty from sequential multi-frame predictions and temporal diversity sampling for non-redundant acquisition.

\section{Method}

\vspace{-0.5mm}
\subsection{Problem Setting}
\label{sec:problem_setting}
\vspace{-0.3mm}
\noindent We formulate \textbf{clip-level active learning} for multi-frame end-to-end MOT, employing fixed-length clips as the acquisition unit.
Let $\mathcal{V}$ denote the training video dataset. 
We build a clip pool $\mathcal{C}$ by extracting temporally ordered length-$T$ clips with a fixed intra-clip interval $\Delta$.
A clip $c\in\mathcal{C}$ is specified by its source video index $\mathrm{vid}(c)\in\mathcal{V}$ and start frame $t_c$.
It is represented as $c=(x_{t_c},x_{t_c+\Delta},\dots,x_{t_c+(T-1)\Delta})$, where $x_t$ is the frame at time $t$.
We denote the sampled frame indices of $c$ as
$\mathcal{K}(c)=\{t_c+k\Delta \mid k=0,\dots,T-1\}$.
Annotating $c$ requires labels for all $T$ frames, so the annotation cost is $T$.
At each round, we maintain a labeled set $\mathcal{L}\subset\mathcal{C}$ and an unlabeled pool $\mathcal{U}=\mathcal{C}\setminus\mathcal{L}$.
Given a per-round budget of $B$ labeled frames, we select a batch $\mathcal{S}_{\mathrm{sel}}\subset\mathcal{U}$ with $|\mathcal{S}_{\mathrm{sel}}|\le b=\lfloor B/T \rfloor$.
For any distinct pair of clips $c_i,c_j\in\mathcal{L}\cup\mathcal{S}_{\mathrm{sel}}$ with $\mathrm{vid}(c_i)=\mathrm{vid}(c_j)$, we require $\mathcal{K}(c_i)\cap\mathcal{K}(c_j)=\emptyset$.

\begin{figure*}[t]
 \centering
 \includegraphics[width=0.90\linewidth]{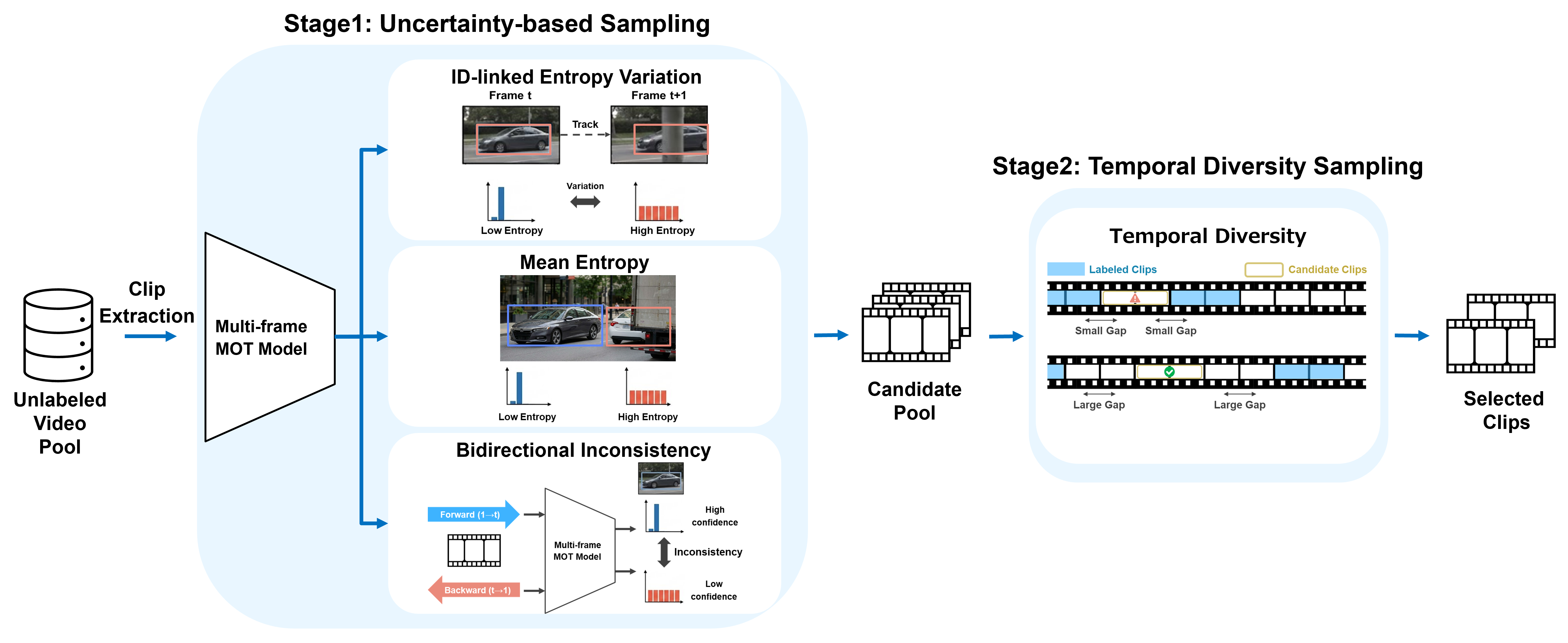}
 \caption{\textbf{Overview of the CUTAL framework.} 
Stage 1: We compute clip uncertainty scores from ID-linked entropy variation, mean entropy, and bidirectional inconsistency, and keep high-uncertainty candidates.
Stage 2: We select a non-redundant batch via temporal diversity sampling based on temporal distances from the labeled set.}
 \label{fig:cutal}
\end{figure*}

\vspace{-0.5mm}
\subsection{Overview of the Proposed Framework}

\vspace{-0.3mm}
\noindent We propose \textbf{CUTAL}, a two-stage framework combining clip uncertainty and temporal diversity (Figure~\ref{fig:cutal}).

\noindent\textbf{Stage 1. Uncertainty-based Sampling.} 
For each $c\in\mathcal{U}$, we compute a clip uncertainty score $S(c)$ from sequential inference results and keep the top $\delta b$ clips as $\mathcal{U}_{\mathrm{cand}}$,
where $b$ is the clip budget and we set $\delta=4$ following \cite{yang2024plug}.

\noindent\textbf{Stage 2. Temporal Diversity Sampling.}
We select $\mathcal{S}_{\mathrm{sel}}$ from $\mathcal{U}_{\mathrm{cand}}$ via k-center greedy \cite{sener2017active,yang2024plug} based on temporal distances within each video.
This spreads annotations over distinct temporal phases of each video while still favoring clips with high uncertainty.

\begin{figure*}[t]
    \begin{adjustbox}{width=\linewidth,center}
    \begin{minipage}[t]{.333\hsize}
        \centering
        \includegraphics[width=\linewidth]{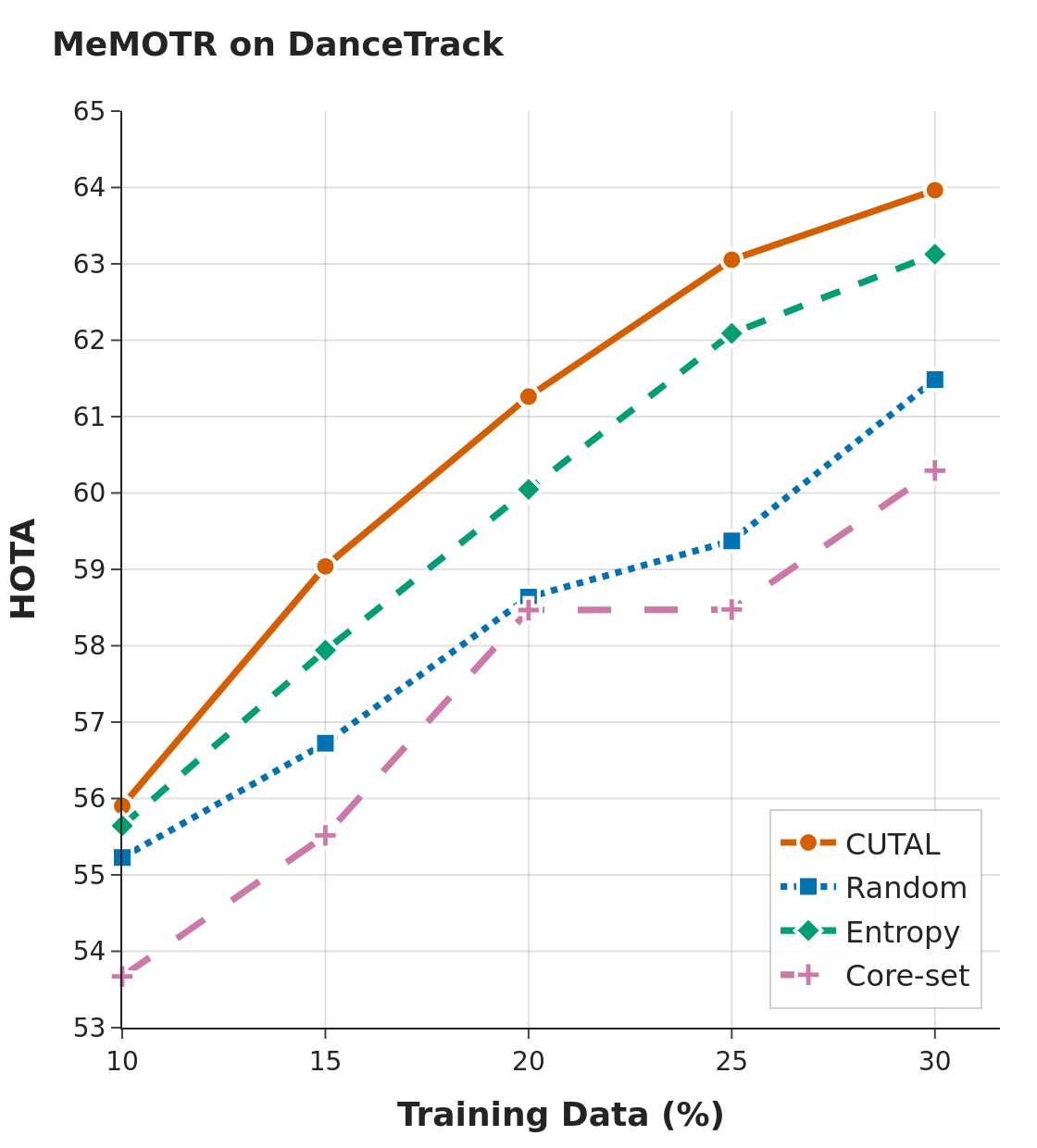}
    \end{minipage}
    \hfill
    \centering

    \centering
    \begin{minipage}[t]{.333\hsize}
        \centering
        \includegraphics[width=\linewidth]{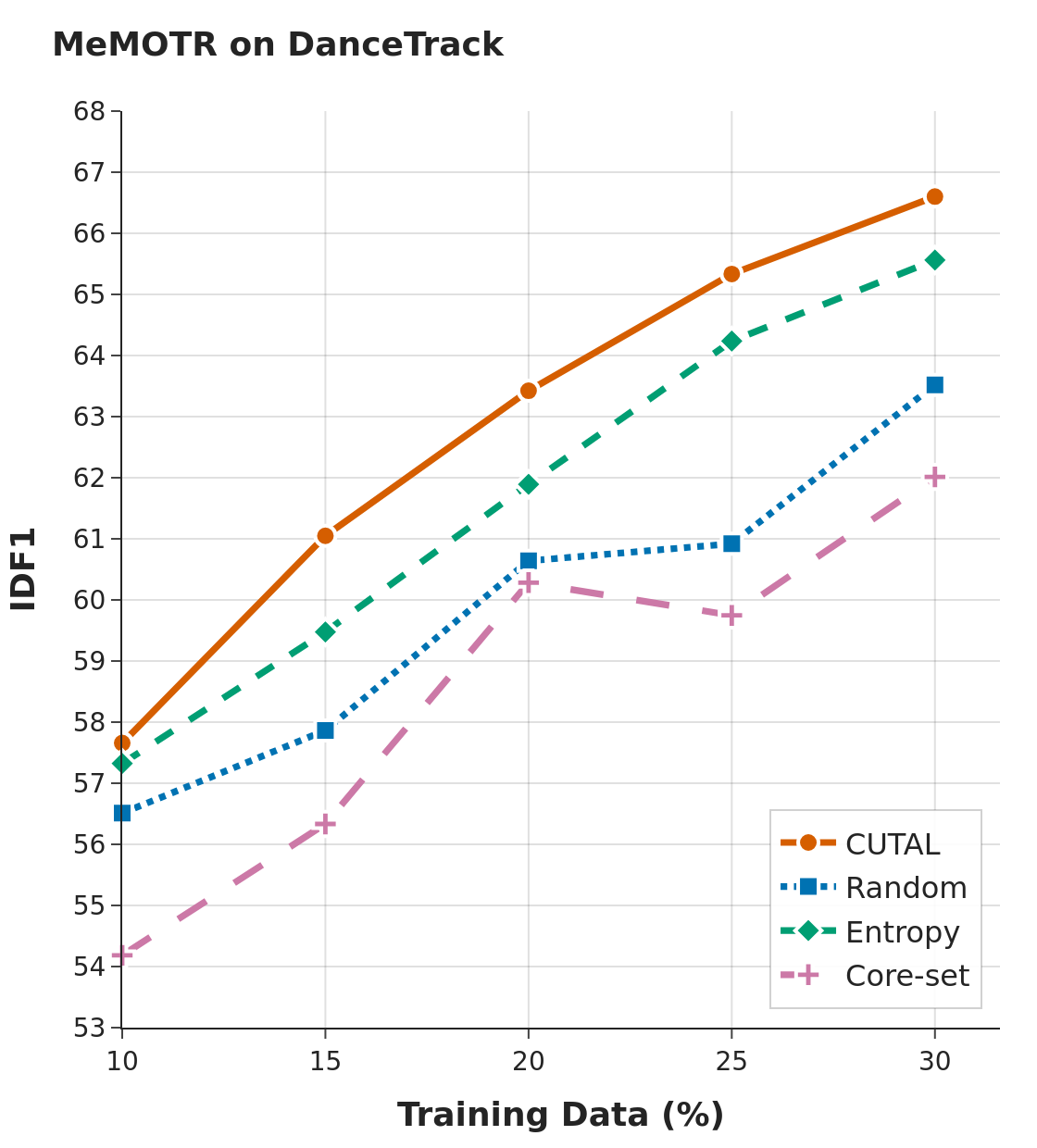}
    \end{minipage}
    \hfill
    \begin{minipage}[t]{.333\hsize}
        \centering
        \includegraphics[width=\linewidth]{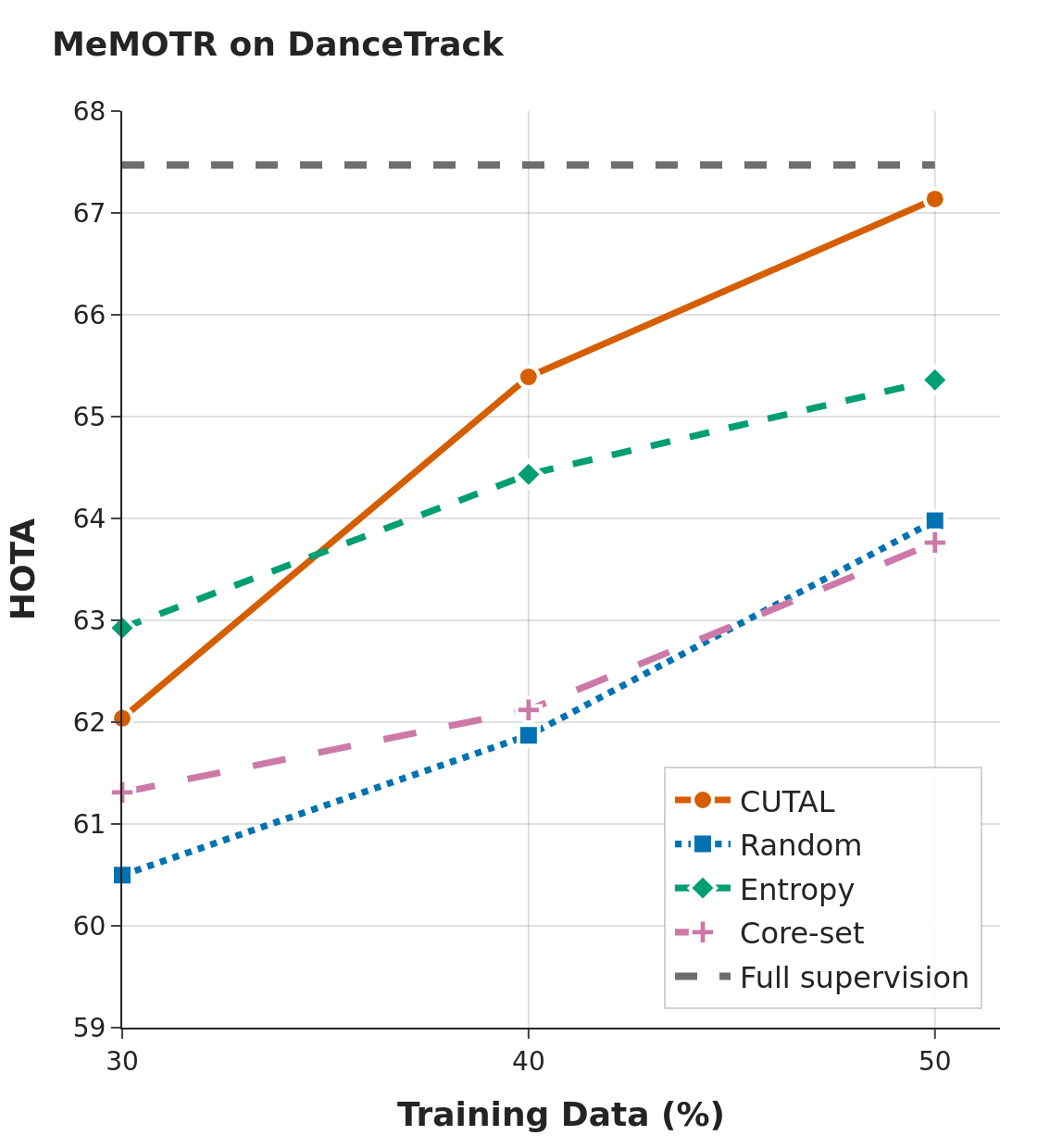}
    \end{minipage}
    \hfill

    \centering
    \begin{minipage}[t]{.333\hsize}
        \centering
        \includegraphics[width=\linewidth]{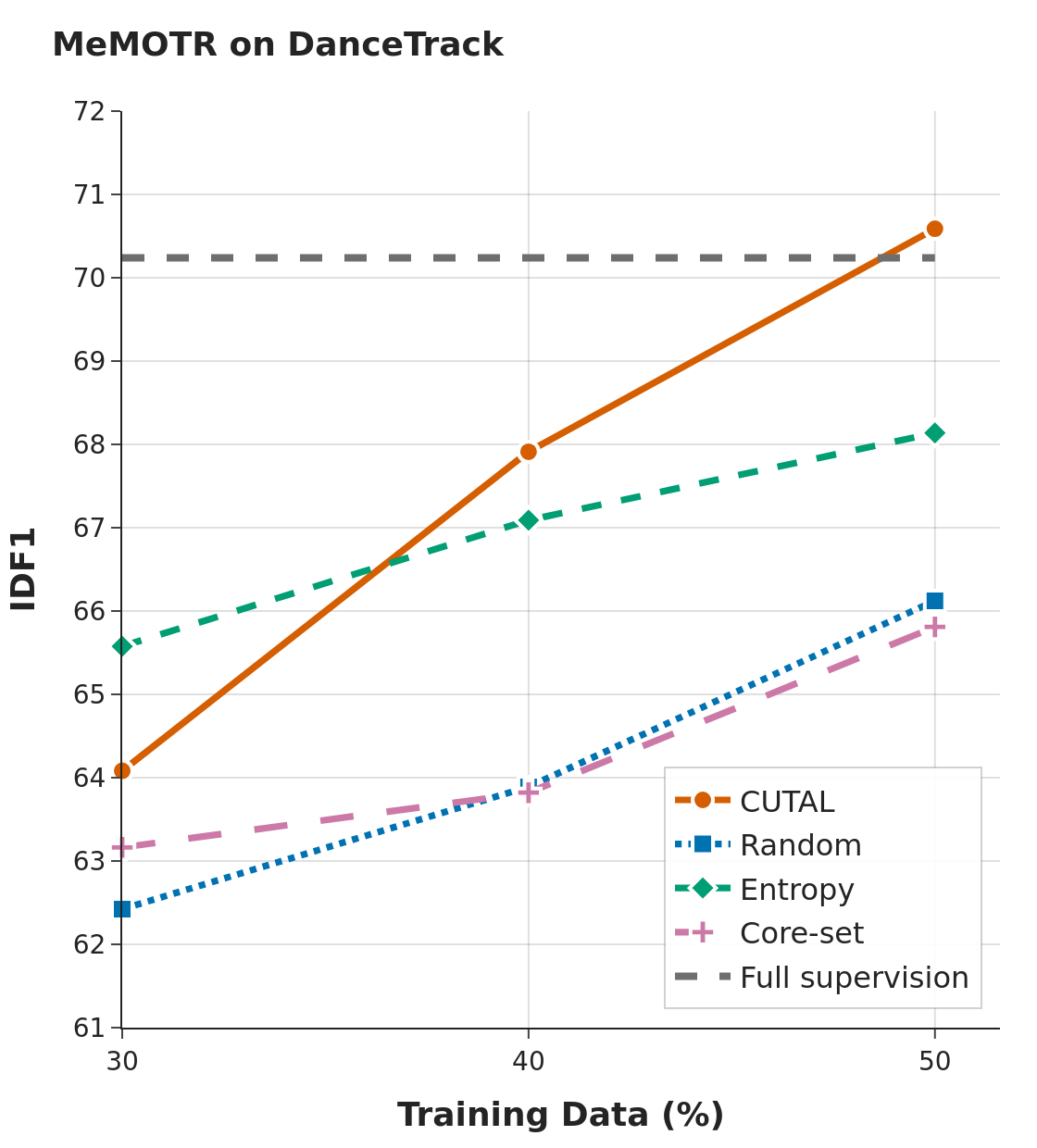}
    \end{minipage}
    \end{adjustbox}
    
    \vspace{-1mm}
    
    \centering
    \begin{adjustbox}{width=\linewidth,center}
    \begin{minipage}[t]{.333\hsize}
        \centering
        \includegraphics[width=\linewidth]{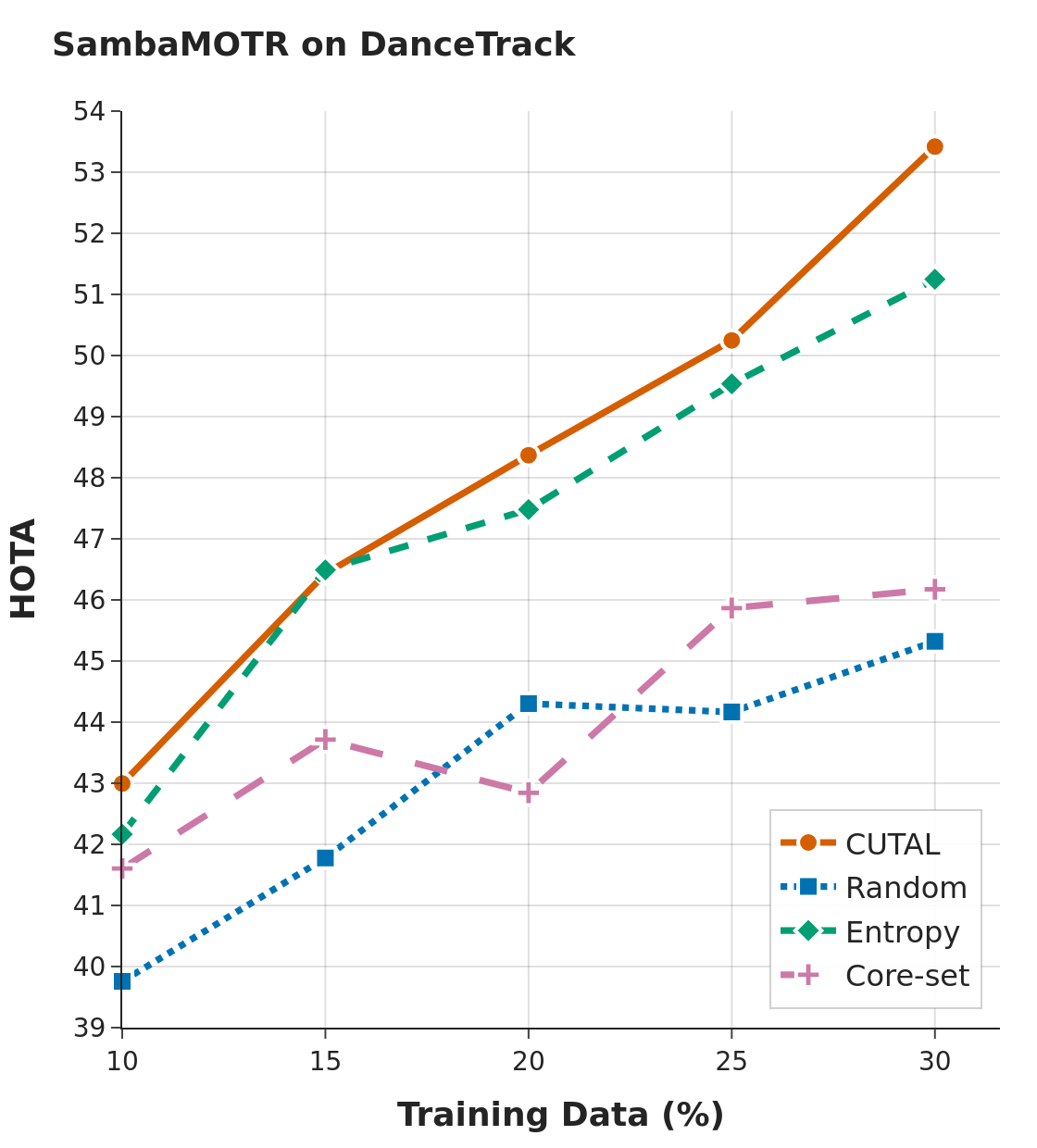}
    \label{fig:55hota}
    \end{minipage}
    \hfill

    \centering
    \begin{minipage}[t]{.333\hsize}
        \centering
        \includegraphics[width=\linewidth]{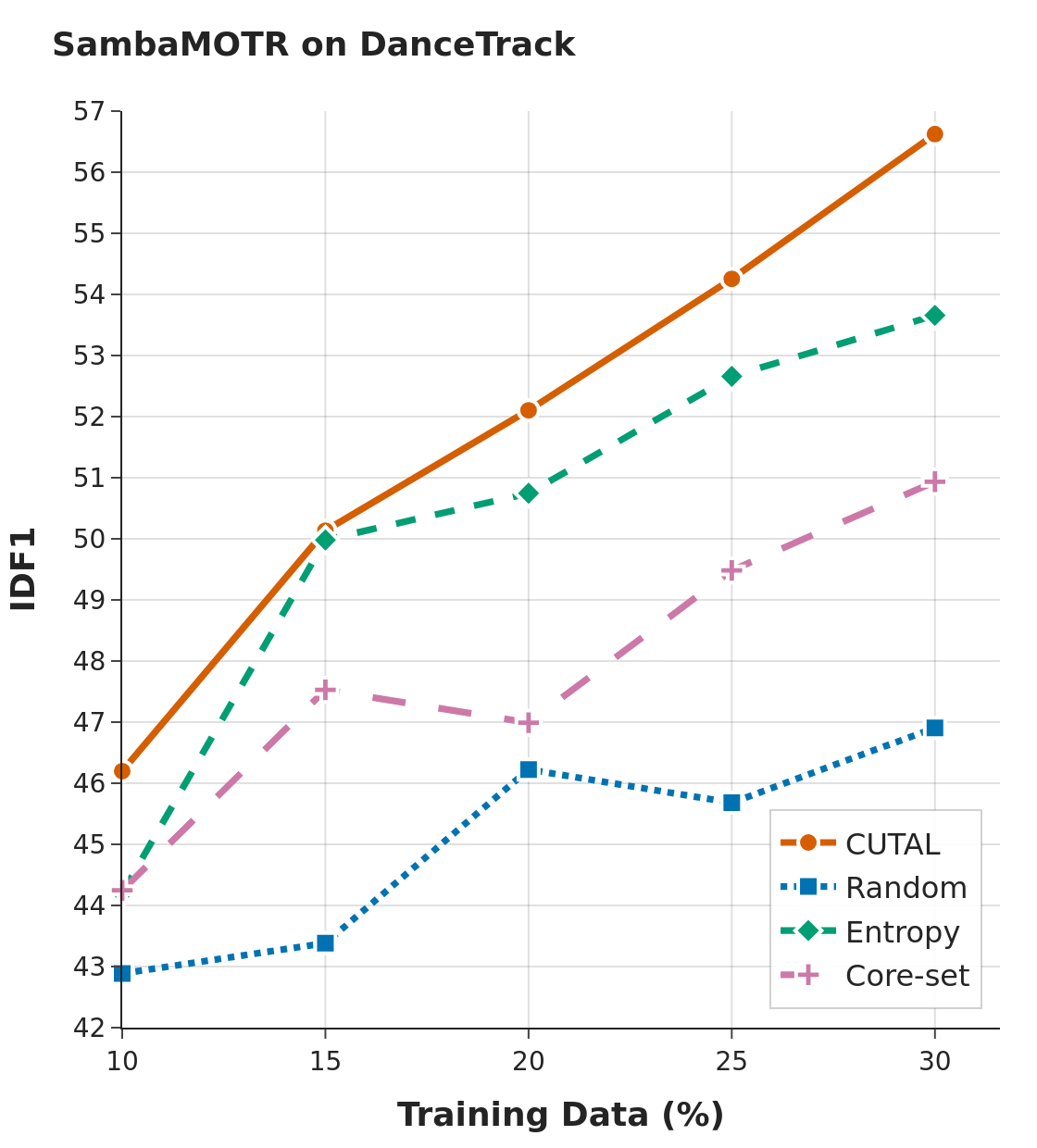}
    \label{fig:55idf1}
    \end{minipage}
    \hfill
    \begin{minipage}[t]{.333\hsize}
        \centering
        \includegraphics[width=\linewidth]{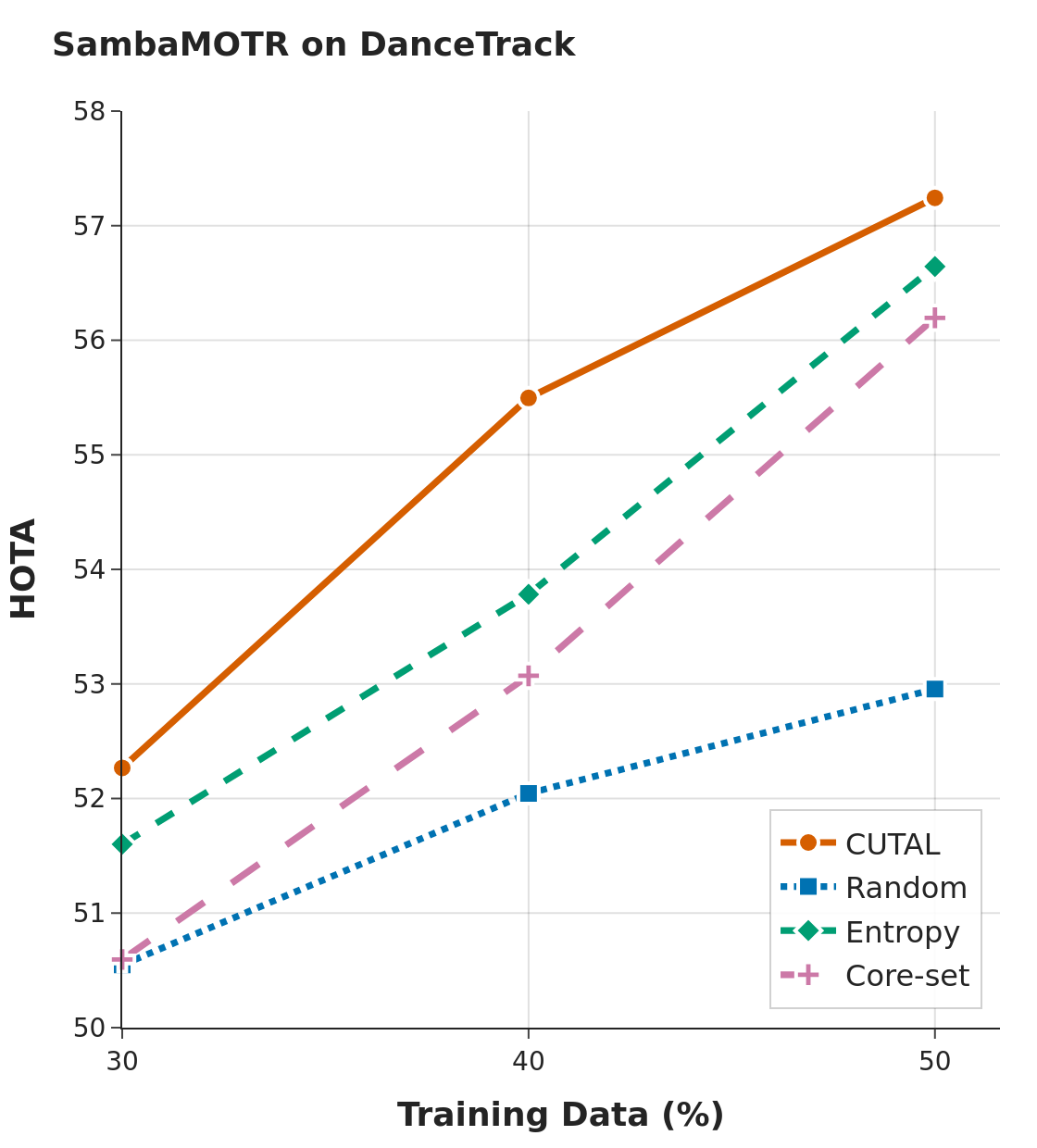}
        \label{fig:2010hota}
    \end{minipage}
    \hfill
    
    \centering
    \begin{minipage}[t]{.333\hsize}
        \centering
        \includegraphics[width=\linewidth]{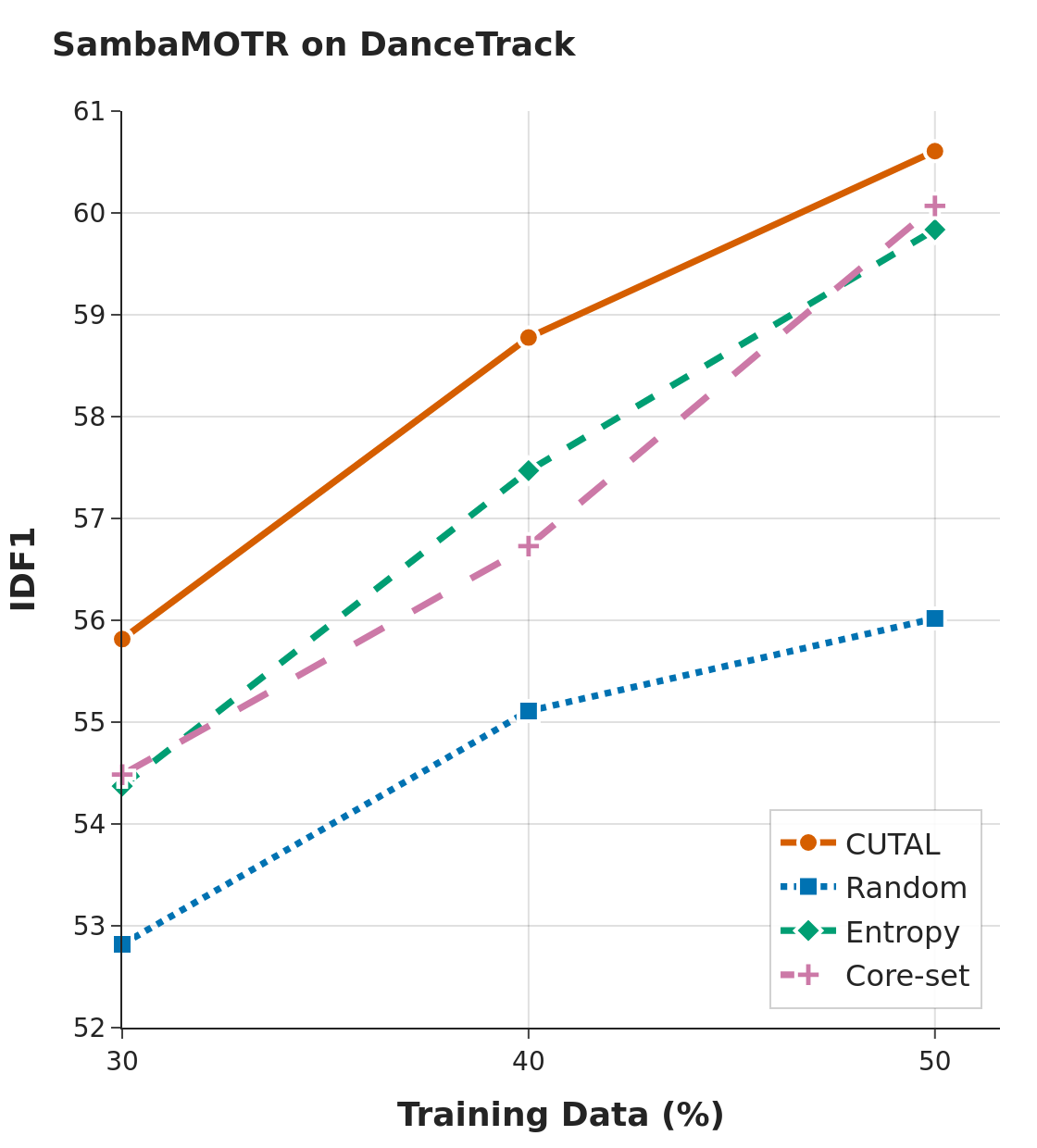}
        \label{fig:2010idf1}
    \end{minipage}
    \end{adjustbox}
    
    \caption{\textbf{Round-wise performance on DanceTrack.}
    We compare HOTA and IDF1 scores of CUTAL against baselines using MeMOTR and SambaMOTR under two budget schedules.
    Note that AssA results are reported in Table~\ref{tab:dancetrack_assa_two_schedules_memotr_sambamotr}.}
    \label{fig:memotr_dancetrack}
\end{figure*}

\vspace{-0.7mm}
\subsection{Clip Uncertainty}
\noindent We quantify uncertainty using three metrics: ID-linked entropy variation $H_{var}(c)$, mean entropy $H_E(c)$, and bidirectional inconsistency $D_{\mathrm{bi}}(c)$.
For a query $q$ at frame $u \in \{0, \dots, T-1\}$, we denote the entropy of its class probability vector $\mathbf{p}_{q,u}$ by $H(\mathbf{p}_{q,u})$, and its confidence score by $s_{q,u}=\max(\mathbf{p}_{q,u})$.

\noindent\textbf{ID-linked Entropy Variation.}
Occlusions and interactions often induce abrupt confidence changes for the same target across consecutive sampled frames.
We measure such instability as the maximum entropy change between ID-matched track queries.
Let $\mathcal{Q}^{\mathrm{trk}}_u$ be the set of active track queries at frame $u$.
For $q \in \mathcal{Q}^{\mathrm{trk}}_u$, let $\hat{\mathrm{id}}_{q,u}$ denote its predicted track identifier.
We form ID-matched query pairs between frames $u$ and $u+1$ as:
{\small
\begin{equation}
\mathcal{P}_u
=
\{(q,q') \mid q\in\mathcal{Q}^{\mathrm{trk}}_u,\ q'\in\mathcal{Q}^{\mathrm{trk}}_{u+1},\ \hat{\mathrm{id}}_{q,u}=\hat{\mathrm{id}}_{q',u+1}\}.
\end{equation}
}
The clip-level variation $H_{var}(c)$ is the average of the maximum entropy change over valid transitions $\mathcal{T}_{var}=\{u \mid |\mathcal{P}_u|>0\}$:
{\small
\begin{equation}
H_{var}(c)
=
\frac{1}{|\mathcal{T}_{var}|}
\sum_{u \in \mathcal{T}_{var}}
\max_{(q, q') \in \mathcal{P}_u}
\bigl|H(\mathbf{p}_{q',u+1}) - H(\mathbf{p}_{q,u})\bigr|.
\end{equation}
}

\noindent\textbf{Mean Entropy.}
While $H_{var}$ captures abrupt temporal fluctuations, it may miss clips where uncertainty persists across frames.
To capture persistent uncertainty, we average over frames the maximum entropy among valid queries at each frame.
Specifically, at each frame $u$, we form a query set $\mathcal{Q}_u$ consisting of active track queries $\mathcal{Q}^{\mathrm{trk}}_u$ and object queries $\mathcal{Q}^{\mathrm{obj}}_u$ whose confidence satisfies $s_{q,u}\ge \tau_{\mathrm{conf}}$ (set to 0.5).
Let $\mathcal{F}_E=\{u \mid |\mathcal{Q}_u|>0\}$ be the set of frames containing valid queries.
We define $H_E(c)$ as:
\begin{equation}
H_E(c)
=
\frac{1}{|\mathcal{F}_E|}
\sum_{u \in \mathcal{F}_E}
\max_{q \in \mathcal{Q}_u}
H(\mathbf{p}_{q,u}).
\end{equation}

\noindent\textbf{Bidirectional Inconsistency.}
While entropy-based metrics quantify classification uncertainty from a single forward pass, reliable tracking also depends on temporal consistency.
We measure this by comparing forward and backward inference, where large discrepancies indicate ambiguous associations or unstable detections.
For each frame $u$, let $\{(b^{\rightarrow}_{i,u}, s^{\rightarrow}_{i,u})\}$ and $\{(b^{\leftarrow}_{j,u}, s^{\leftarrow}_{j,u})\}$ be the bounding boxes and scores from bidirectional inference.
We match these predictions via greedy IoU matching (threshold $\tau_{\mathrm{IoU}}=0.5$) to form pairs $\mathcal{M}_u$.
Averaging the maximum score difference over matched frames $\mathcal{F}_{bi}=\{u \mid |\mathcal{M}_u|>0\}$ yields the bidirectional inconsistency score $D_{\mathrm{bi}}(c)$:
\begin{equation}
D_{\mathrm{bi}}(c)
=
\frac{1}{|\mathcal{F}_{bi}|} \sum_{u \in \mathcal{F}_{bi}}
\max_{(i,j)\in\mathcal{M}_u} \bigl|s^{\rightarrow}_{i,u} - s^{\leftarrow}_{j,u}\bigr|.
\end{equation}

\noindent\textbf{Score Aggregation.}
To combine these metrics, we normalize each component using its mean $\mu$ and standard deviation $\sigma$ computed over the unlabeled pool $\mathcal{U}$.
Following \cite{nakamura2024active}, we use the normalization:

\begin{equation}
\phi(x)
=
\max\!\left(
0,\,
\frac{x - (\mu - 3\sigma)}{6\sigma}
\right).
\end{equation}
The final uncertainty score is the product of these normalized values:
\begin{equation}
S(c)
=
\phi\bigl(H_E(c)\bigr) \cdot \phi\bigl(H_{var}(c)\bigr) \cdot \phi\bigl(D_{\mathrm{bi}}(c)\bigr).
\end{equation}

\subsection{Temporal Diversity Sampling}
\noindent Relying solely on uncertainty scores leads to redundancy, as high uncertainty often persists across adjacent clips.
To mitigate this, we diversify selected clips over the timeline of each video.

For two clips $c_i$ and $c_j$ from the same video, we define the temporal gap as:
\begin{equation}
\mathrm{gap}(c_i,c_j)=\max\!\bigl(0,\,|t_{c_i}-t_{c_j}|-(T-1)\Delta\bigr).
\end{equation}
Our selection strategy employs a k-center greedy approach to ensure temporal coverage.
Let $\mathcal{S}_{\mathrm{sel}}$ denote the set of clips selected so far in the current round.
For a candidate clip $c$, we compute the minimum temporal distance to the annotated set $\mathcal{L}\cup\mathcal{S}_{\mathrm{sel}}$ within the same video:
\begin{equation}
d_{\mathrm{near}}(c)
=
\min_{c' \in \mathcal{L} \cup \mathcal{S}_{\mathrm{sel}},\, \mathrm{vid}(c') = \mathrm{vid}(c)}
\mathrm{gap}(c,c').
\end{equation}
The selection process iteratively adds the candidate that maximizes $d_{\mathrm{near}}(c)$.
This spreads annotations across the timeline and avoids selecting temporally redundant clips under the fixed budget.

\section{Experiments}
\label{sec:experiments}
\noindent We evaluate CUTAL to verify its effectiveness for clip-level active learning in multi-object tracking.
Due to space constraints, we report the quantitative comparisons in this section.
Qualitative results, ablation studies, discussions, and the complete tabular data for the curves in Figure~\ref{fig:memotr_dancetrack} are provided in the supplementary materials. 

\subsection{Datasets and Evaluation Metrics}
\noindent We evaluate CUTAL on two benchmarks, DanceTrack \cite{sun2022dancetrack} and SportsMOT \cite{cui2023sportsmot}.
For both datasets, we use the training split as the unlabeled pool.
We report results on the DanceTrack test set via its public evaluation server.
For SportsMOT, given the high evaluation frequency required for our multi-round active learning experiments, we report performance on the validation split since the official test server has a limited submission frequency.
We use HOTA \cite{luiten2021hota} as the main MOT metric and additionally report AssA and IDF1 \cite{ristani2016performance} to measure association quality.
Results are averaged over three random seeds.

\begin{table}[t]
\centering
\caption{\textbf{Round-wise AssA results on DanceTrack under two budget schedules.}
Best results are marked in \textbf{bold} and second best are \underline{underlined}.}
\label{tab:dancetrack_assa_two_schedules_memotr_sambamotr}
\vspace{1.3mm}
\scriptsize
\setlength{\tabcolsep}{3pt}
\renewcommand{\arraystretch}{1.25}

\begin{adjustbox}{width=0.93\linewidth,center}
\begin{tabular}{l | ccccc | ccc}
\hline

Method
& \multicolumn{5}{c|}{5\%+5\%}
& \multicolumn{3}{c}{20\%+10\%} \\
\cline{2-9}

& 10\% & 15\% & 20\% & 25\% & 30\%
& 30\% & 40\% & 50\% \\
\hline

\multicolumn{9}{l}{\textbf{MeMOTR}} \\
Random
& 40.87 & 42.56 & 45.43 & 45.66 & 48.29
& 47.23 & 49.20 & 51.90 \\
Entropy
& \underline{42.04} & \underline{44.53} & \underline{46.95} & \underline{49.52} & \underline{50.93}
& \textbf{50.65} & \underline{52.48} & \underline{53.94} \\
Core-set
& 38.92 & 40.96 & 44.62 & 44.83 & 46.77
& 48.20 & 49.39 & 51.60 \\
CUTAL
& \textbf{42.10} & \textbf{45.51} & \textbf{48.41} & \textbf{50.91} & \textbf{52.22}
& \underline{49.36} & \textbf{53.99} & \textbf{56.63} \\
\cline{1-9}
\textit{Full supervision}
& \multicolumn{8}{c}{56.87} \\
\hline
\hline

\multicolumn{9}{l}{\textbf{SambaMOTR}} \\
Random
& 28.10 & 29.14 & 31.60 & 31.30 & 32.66
& 38.71 & 39.70 & 41.65 \\
Entropy
& \underline{29.82} & \underline{34.91} & \underline{36.03} & \underline{38.25} & \underline{39.02}
& 39.67 & \underline{42.25} & 45.19 \\
Core-set
& 29.06 & 33.01 & 32.57 & 34.26 & 36.44
& \underline{39.85} & 41.45 & \underline{45.27} \\
CUTAL
& \textbf{32.12} & \textbf{35.44} & \textbf{37.99} & \textbf{40.07} & \textbf{41.95}
& \textbf{41.30} & \textbf{44.23} & \textbf{46.06} \\

\cline{1-9}
\textit{Full supervision}
& \multicolumn{8}{c}{55.55} \\
\hline

\end{tabular}
\end{adjustbox}
\end{table}

\begin{table*}[t]
\centering
\caption{\textbf{Round-wise results on SportsMOT under the 5\%+5\% schedule.}
}
\label{tab:sportsmot_5p5p_memotr_sambamotr_vertical}
\vspace{1.3mm}
\scriptsize
\setlength{\tabcolsep}{3pt}
\renewcommand{\arraystretch}{1.25}

\begin{adjustbox}{width=0.92\linewidth,center}
\begin{tabular}{l | ccccc | ccccc | ccccc}
\hline
Method
& \multicolumn{5}{c|}{HOTA}
& \multicolumn{5}{c|}{AssA}
& \multicolumn{5}{c}{IDF1} \\
\cline{2-16}
& 10\% & 15\% & 20\% & 25\% & 30\%
& 10\% & 15\% & 20\% & 25\% & 30\%
& 10\% & 15\% & 20\% & 25\% & 30\% \\
\hline

\multicolumn{16}{l}{\textbf{MeMOTR}} \\
Random
& 68.38 & 69.61 & \underline{71.60} & 71.78 & \underline{72.31}
& 57.21 & 58.54 & \underline{61.19} & 60.98 & 62.04
& 70.65 & 70.92 & \underline{74.41} & 74.08 & 74.83 \\
Entropy
& \underline{68.61} & \underline{69.95} & 70.73 & 72.24 & 72.14
& \underline{57.50} & \underline{58.27} & 60.93 & 62.05 & \underline{62.29}
& \underline{71.49} & \underline{71.94} & 73.70 & 73.70 & \underline{75.47} \\
Core-set
& 67.52 & 68.43 & 70.78 & \underline{72.50} & 71.95
& 56.60 & 57.03 & 60.04 & \underline{62.25} & 61.76
& 70.67 & 70.80 & 73.19 & \underline{75.21} & 74.42 \\
CUTAL
& \textbf{69.35} & \textbf{71.06} & \textbf{71.82} & \textbf{72.92} & \textbf{73.56}
& \textbf{57.91} & \textbf{60.41} & \textbf{61.70} & \textbf{63.04} & \textbf{64.17}
& \textbf{71.59} & \textbf{73.75} & \textbf{74.93} & \textbf{76.02} & \textbf{76.83} \\

\hline
\hline

\multicolumn{16}{l}{\textbf{SambaMOTR}} \\
Random
& 55.52 & 56.53 & 58.27 & 57.23 & 56.89
& 45.82 & 45.31 & 47.16 & 45.93 & 46.47
& 58.77 & 58.65 & 60.81 & 59.72 & 59.69 \\
Entropy
& \textbf{57.39} & \underline{58.17} & 59.52 & \underline{62.84} & \underline{62.96}
& \textbf{47.99} & \underline{48.47} & 49.52 & \underline{53.11} & \underline{53.25}
& \textbf{61.40} & \underline{61.82} & 62.97 & \underline{66.69} & \underline{66.65} \\
Core-set
& 56.42 & 56.20 & \underline{60.11} & 60.64 & 61.72
& 46.75 & 48.25 & \underline{50.16} & 50.40 & 52.23
& 59.96 & 60.53 & \underline{63.73} & 63.90 & 65.13 \\
CUTAL
& \underline{56.65} & \textbf{58.69} & \textbf{62.13} & \textbf{63.84} & \textbf{64.54}
& \underline{46.84} & \textbf{48.59} & \textbf{51.91} & \textbf{53.69} & \textbf{54.16}
& \underline{59.94} & \textbf{62.40} & \textbf{65.68} & \textbf{67.49} & \textbf{67.88} \\

\hline

\end{tabular}
\end{adjustbox}
\end{table*}

\begin{table}[t]
\centering
\caption{\textbf{Round-wise results on SportsMOT under the 20\%+10\% schedule.}
}
\label{tab:sportsmot_20p10_memotr_sambamotr_vertical}
\vspace{1.3mm}
\scriptsize
\setlength{\tabcolsep}{3.5pt}
\renewcommand{\arraystretch}{1.25}

\begin{adjustbox}{width=0.94\linewidth,center}
\begin{tabular}{l | ccc | ccc | ccc}
\hline
Method
& \multicolumn{3}{c|}{HOTA}
& \multicolumn{3}{c|}{AssA}
& \multicolumn{3}{c}{IDF1} \\
\cline{2-10}
& 30\% & 40\% & 50\%
& 30\% & 40\% & 50\%
& 30\% & 40\% & 50\% \\
\hline

\multicolumn{10}{l}{\textbf{MeMOTR}} \\
Random
& 71.81 & 73.22 & 73.77
& 61.49 & 63.28 & 64.48
& 74.48 & 75.98 & 76.71 \\
Entropy
& 71.66 & \underline{73.33} & \underline{74.01}
& 61.17 & \underline{63.54} & \underline{64.86}
& 74.33 & \underline{76.34} & \underline{77.13} \\
Core-set
& \textbf{72.73} & 73.18 & 73.64
& \textbf{62.44} & 63.39 & 64.17
& \textbf{75.31} & 76.02 & 76.84 \\
CUTAL
& \underline{72.29} & \textbf{74.19} & \textbf{74.23}
& \underline{62.28} & \textbf{65.21} & \textbf{65.17}
& \underline{75.25} & \textbf{77.69} & \textbf{77.72} \\
\cline{1-10}

\textit{Full supervision}
& \multicolumn{3}{c|}{74.47}
& \multicolumn{3}{c|}{65.54}
& \multicolumn{3}{c}{77.47} \\
\hline
\hline

\multicolumn{10}{l}{\textbf{SambaMOTR}} \\
Random
& 62.27 & 64.08 & 63.55
& 52.87 & 54.10 & 53.37
& 65.98 & 67.58 & 66.76 \\
Entropy
& \textbf{64.18} & 62.77 & 67.38
& \textbf{54.16} & \underline{53.56} & 58.27
& \textbf{67.61} & 66.60 & 71.29 \\
Core-set
& 61.51 & \underline{63.37} & \underline{67.41}
& 52.99 & 53.54 & \underline{57.46}
& 65.82 & \underline{66.86} & \underline{70.67} \\
CUTAL
& \underline{62.28} & \textbf{66.31} & \textbf{68.88}
& \underline{52.36} & \textbf{56.27} & \textbf{59.70}
& \underline{65.82} & \textbf{69.77} & \textbf{72.65} \\

\cline{1-10}
\textit{Full supervision}
& \multicolumn{3}{c|}{77.27}
& \multicolumn{3}{c|}{69.51}
& \multicolumn{3}{c}{80.96} \\
\hline
\end{tabular}
\end{adjustbox}
\end{table}

\subsection{Implementation Details}
\noindent\textbf{Models and Training.}
We employ MeMOTR \cite{gao2023memotr} and SambaMOTR \cite{segu2025samba} as representative Transformer-based end-to-end MOT models.
For both architectures, we follow the original network designs and loss functions.
We set $T$ to values close to the temporal span used in the original training recipes of each tracker.
Specifically, we use $T=4$ for MeMOTR and $T=10$ for SambaMOTR.
To build the clip set $\mathcal{C}$ from each training video, we sample $T$ frames with a fixed intra-clip interval $\Delta=5$ in all experiments.
We follow the official configurations for data augmentation, learning rate schedules, and other hyperparameters.

\noindent\textbf{Active Learning Protocol.}
The budget $B$ is a percentage of the total training dataset frames, and the number of selected clips $b$ follows Section~\ref{sec:problem_setting}.
We use two schedules: (1) start at 5\% and add 5\% per round (5\%+5\%), and (2) start at 20\% and add 10\% per round (20\%+10\%).

\subsection{Baselines} 
\noindent We compare CUTAL against the following baselines:
\noindent\textbf{Random.}
Selects clips uniformly at random from the unlabeled pool.
\noindent\textbf{Entropy.}
Computes the entropy of the predicted class scores for each track at each frame, takes the maximum value over tracks for each frame, and averages it over all frames in the clip.
\noindent\textbf{Core-set.}
Performs k-center greedy selection in the clip feature space.
The clip feature is obtained by averaging track query embeddings from the final transformer decoder layer across all frames and tracks in the clip.

\subsection{Quantitative Results}

\noindent\textbf{DanceTrack.}
Figure~\ref{fig:memotr_dancetrack} and Table~\ref{tab:dancetrack_assa_two_schedules_memotr_sambamotr} report the results on DanceTrack.
For MeMOTR, CUTAL achieves the best overall performance under the 5\%+5\% schedule.
Under the 20\%+10\% schedule, Entropy temporarily leads at 30\% labels, which may be attributed to the cold-start effect \cite{chen2024making}, where acquisition scores can be noisy when the labeled set is still limited.
As more labels are acquired, CUTAL achieves the best performance in later rounds.
At 50\% labels, CUTAL reaches 67.14 HOTA, closely approaching full supervision at 67.47 HOTA.
For SambaMOTR, CUTAL yields larger gains, especially in association quality.
Under the 5\%+5\% schedule, CUTAL leads AssA and IDF1 in every round and reaches 53.42 HOTA at 30\% labels, improving over Entropy by 2.17.
Under the 20\%+10\% schedule, CUTAL continues to perform best from 30\% to 50\% labels.

\noindent\textbf{SportsMOT.}
Table~\ref{tab:sportsmot_5p5p_memotr_sambamotr_vertical} and Table~\ref{tab:sportsmot_20p10_memotr_sambamotr_vertical} report the results on SportsMOT.
For MeMOTR, CUTAL consistently achieves the best performance under the 5\%+5\% schedule.
Under the 20\%+10\% schedule, it overtakes Core-set after 30\% labels and is on par with full supervision at 50\% labels, reaching 74.23 HOTA and 65.17 AssA, while slightly improving IDF1.
For SambaMOTR, CUTAL becomes the best method from 15\% labels onward under the 5\%+5\% schedule.
Under the 20\%+10\% schedule, CUTAL outperforms all baselines at 40\% and 50\% labels.
Specifically, at 50\% labels, CUTAL reaches 68.88 HOTA, 59.70 AssA, and 72.65 IDF1, demonstrating its consistent superiority in both detection and association stability.
\label{sec:exp}
\vspace{-5mm}
\section{Conclusion}
\label{sec:conclusion}
\noindent We presented CUTAL, a clip-level active learning framework for Transformer-based end-to-end MOT.
By shifting the acquisition unit from isolated frames to temporally ordered clips and integrating sequential uncertainty with temporal diversity sampling, CUTAL addresses the structural mismatch in frame-level acquisition.
Experiments on DanceTrack and SportsMOT show that CUTAL provides stronger overall performance than baselines under matched annotation budgets.
Notably, CUTAL enables MeMOTR to achieve near full-supervision performance using only 50\% labeled data.
{
    \small
    \putbib[main]
}
\end{bibunit}


\clearpage

\setcounter{section}{0}
\setcounter{subsection}{0}
\setcounter{subsubsection}{0}
\setcounter{figure}{0}
\setcounter{table}{0}
\setcounter{equation}{0}

\renewcommand{\thesection}{S\arabic{section}}
\renewcommand{\thesubsection}{S\arabic{section}.\arabic{subsection}}
\renewcommand{\thesubsubsection}{S\arabic{section}.\arabic{subsection}.\arabic{subsubsection}}
\renewcommand{\thefigure}{S\arabic{figure}}
\renewcommand{\thetable}{S\arabic{table}}
\renewcommand{\theequation}{S\arabic{equation}}

\twocolumn[{
\begin{center}
{\large\bf
SUPPLEMENTARY MATERIAL:\par
\vspace{0.5em}
CLIP-LEVEL UNCERTAINTY AND TEMPORAL-AWARE ACTIVE LEARNING FOR\par
END-TO-END MULTI-OBJECT TRACKING\par
}
\vspace{1.5em}
{\large
\textit{Riku Inoue, \qquad Shogo Sato, \qquad Kazuhiko Murasaki, \qquad Tomoyasu Shimada,}\par
\textit{Toshihiko Nishimura, \qquad Ryuichi Tanida}\par
}
\vspace{1.0em}
{\large NTT, Inc., Kanagawa, Japan\par}
\end{center}
\vspace{1.7em}
}]

\begin{bibunit}[icip]
\newcommand{\subcaptext}[1]{\vspace{-1mm}\par\centering{\small #1}\par}

\section{Introduction}
\label{sec:supp_intro}

\noindent This supplementary material provides additional results and analyses supporting the main paper.
We include complete round-wise results on DanceTrack~\cite{sun2022dancetrack} under two budget schedules, qualitative comparisons, ablations, sampling visualizations, and discussions.

\section{Additional Quantitative Results on DanceTrack}
\label{sec:supp_quant_dancetrack}

\noindent Table~\ref{tab:dancetrack_5p5p_memotr_sambamotr_vertical} and Table~\ref{tab:dancetrack_20p10_memotr_sambamotr_vertical} report the complete round-wise results on DanceTrack \cite{sun2022dancetrack} for MeMOTR \cite{gao2023memotr} and SambaMOTR \cite{segu2025samba}.
We report HOTA, AssA, and IDF1, corresponding to Figure 3 in the main paper.

\begin{table*}[t]
\centering
\caption{\textbf{Round-wise results on DanceTrack under the 5\%+5\% schedule.}
Best results are marked in \textbf{bold} and second best are \underline{underlined}.}
\label{tab:dancetrack_5p5p_memotr_sambamotr_vertical}
\vspace{1.3mm}
\scriptsize

\setlength{\tabcolsep}{2.5pt}
\renewcommand{\arraystretch}{1.25}

\begin{adjustbox}{width=0.9\linewidth,center}
\begin{tabular}{l | ccccc | ccccc | ccccc}
\hline
Method
& \multicolumn{5}{c|}{HOTA}
& \multicolumn{5}{c|}{AssA}
& \multicolumn{5}{c}{IDF1} \\
\cline{2-16}
& 10\% & 15\% & 20\% & 25\% & 30\%
& 10\% & 15\% & 20\% & 25\% & 30\%
& 10\% & 15\% & 20\% & 25\% & 30\% \\
\hline

\multicolumn{16}{l}{\textbf{MeMOTR}} \\
Random
& 55.23 & 56.72 & 58.64 & 59.37 & 61.48
& 40.87 & 42.56 & 45.43 & 45.66 & 48.29
& 56.51 & 57.86 & 60.64 & 60.92 & 63.52 \\
Entropy
& \underline{55.64} & \underline{57.94} & \underline{60.05} & \underline{62.09} & \underline{63.13}
& \underline{42.04} & \underline{44.53} & \underline{46.95} & \underline{49.52} & \underline{50.93}
& \underline{57.32} & \underline{59.48} & \underline{61.89} & \underline{64.24} & \underline{65.56} \\
Core-set
& 53.67 & 55.52 & 58.47 & 58.47 & 60.29
& 38.92 & 40.96 & 44.62 & 44.83 & 46.77
& 54.18 & 56.33 & 60.28 & 59.75 & 62.01 \\
CUTAL
& \textbf{55.90} & \textbf{59.04} & \textbf{61.26} & \textbf{63.05} & \textbf{63.96}
& \textbf{42.10} & \textbf{45.51} & \textbf{48.41} & \textbf{50.91} & \textbf{52.22}
& \textbf{57.66} & \textbf{61.05} & \textbf{63.42} & \textbf{65.33} & \textbf{66.60} \\
\hline
\hline

\multicolumn{16}{l}{\textbf{SambaMOTR}} \\
Random
& 39.76 & 41.78 & 44.30 & 44.17 & 45.32
& 28.10 & 29.14 & 31.60 & 31.30 & 32.66
& 42.89 & 43.38 & 46.22 & 45.68 & 46.91 \\
Entropy
& \underline{42.17} & \textbf{46.49} & \underline{47.48} & \underline{49.54} & \underline{51.25}
& \underline{29.82} & \underline{34.91} & \underline{36.03} & \underline{38.25} & \underline{39.02}
& 44.20 & \underline{49.98} & \underline{50.75} & \underline{52.66} & \underline{53.66} \\
Core-set
& 41.60 & 43.72 & 42.84 & 45.87 & 46.17
& 29.06 & 33.01 & 32.57 & 34.26 & 36.44
& \underline{44.25} & 47.53 & 46.99 & 49.48 & 50.94 \\
CUTAL
& \textbf{42.99} & \underline{46.43} & \textbf{48.37} & \textbf{50.25} & \textbf{53.42}
& \textbf{32.12} & \textbf{35.44} & \textbf{37.99} & \textbf{40.07} & \textbf{41.95}
& \textbf{46.20} & \textbf{50.13} & \textbf{52.10} & \textbf{54.25} & \textbf{56.62} \\
\hline
\end{tabular}
\end{adjustbox}
\end{table*}

\begin{table}[t]
\centering
\caption{\textbf{Round-wise results on DanceTrack under the 20\%+10\% schedule.}}
\label{tab:dancetrack_20p10_memotr_sambamotr_vertical}
\vspace{1.3mm}
\scriptsize
\setlength{\tabcolsep}{3.5pt}
\renewcommand{\arraystretch}{1.25}

\begin{adjustbox}{width=0.98\linewidth,center}
\begin{tabular}{l | ccc | ccc | ccc}
\hline
Method
& \multicolumn{3}{c|}{HOTA}
& \multicolumn{3}{c|}{AssA}
& \multicolumn{3}{c}{IDF1} \\
\cline{2-10}
& 30\% & 40\% & 50\%
& 30\% & 40\% & 50\%
& 30\% & 40\% & 50\% \\
\hline

\multicolumn{10}{l}{\textbf{MeMOTR}} \\
Random
& 60.50 & 61.87 & 63.98
& 47.23 & 49.20 & 51.90
& 62.42 & 63.89 & 66.12 \\
Entropy
& \textbf{62.92} & \underline{64.43} & \underline{65.36}
& \textbf{50.65} & \underline{52.48} & \underline{53.94}
& \textbf{65.58} & \underline{67.09} & \underline{68.14} \\
Core-set
& 61.31 & 62.12 & 63.76
& 48.20 & 49.39 & 51.60
& 63.16 & 63.82 & 65.81 \\
CUTAL
& \underline{62.04} & \textbf{65.39} & \textbf{67.14}
& \underline{49.36} & \textbf{53.99} & \textbf{56.63}
& \underline{64.08} & \textbf{67.91} & \textbf{70.59} \\
\cline{1-10}

\textit{Full supervision}
& \multicolumn{3}{c|}{67.47}
& \multicolumn{3}{c|}{56.87}
& \multicolumn{3}{c}{70.24} \\
\hline
\hline

\multicolumn{10}{l}{\textbf{SambaMOTR}} \\
Random
& 50.55 & 52.05 & 52.96
& 38.71 & 39.70 & 41.65
& 52.82 & 55.11 & 56.02 \\
Entropy
& \underline{51.60} & \underline{53.78} & \underline{56.64}
& 39.67 & \underline{42.25} & 45.19
& 54.37 & \underline{57.47} & 59.84 \\
Core-set
& 50.59 & 53.07 & 56.20
& \underline{39.85} & 41.45 & \underline{45.27}
& \underline{54.48} & 56.73 & \underline{60.07} \\
CUTAL
& \textbf{52.27} & \textbf{55.50} & \textbf{57.24}
& \textbf{41.30} & \textbf{44.23} & \textbf{46.06}
& \textbf{55.82} & \textbf{58.78} & \textbf{60.61} \\
\cline{1-10}

\textit{Full supervision}
& \multicolumn{3}{c|}{66.28}
& \multicolumn{3}{c|}{55.55}
& \multicolumn{3}{c}{69.84} \\
\hline
\end{tabular}
\end{adjustbox}
\end{table}

\section{Qualitative Results}
\label{sec:supp_qualitative}

\begin{figure*}[t]
    \centering
    \newcommand{\figwidth}{0.33\linewidth}

    \begin{minipage}[t]{\linewidth}
        \centering
        \includegraphics[width=\figwidth]{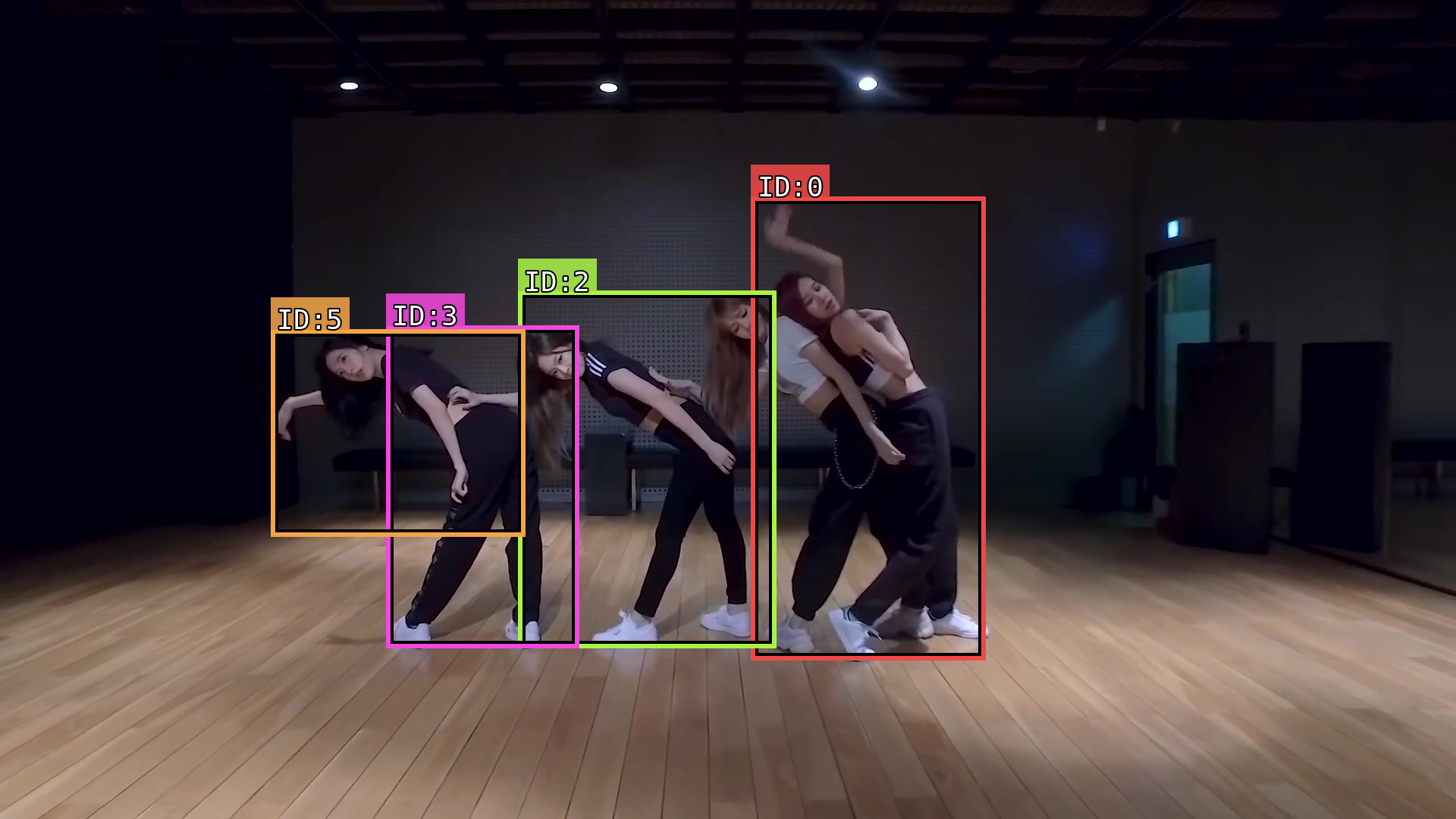}\hfill
        \includegraphics[width=\figwidth]{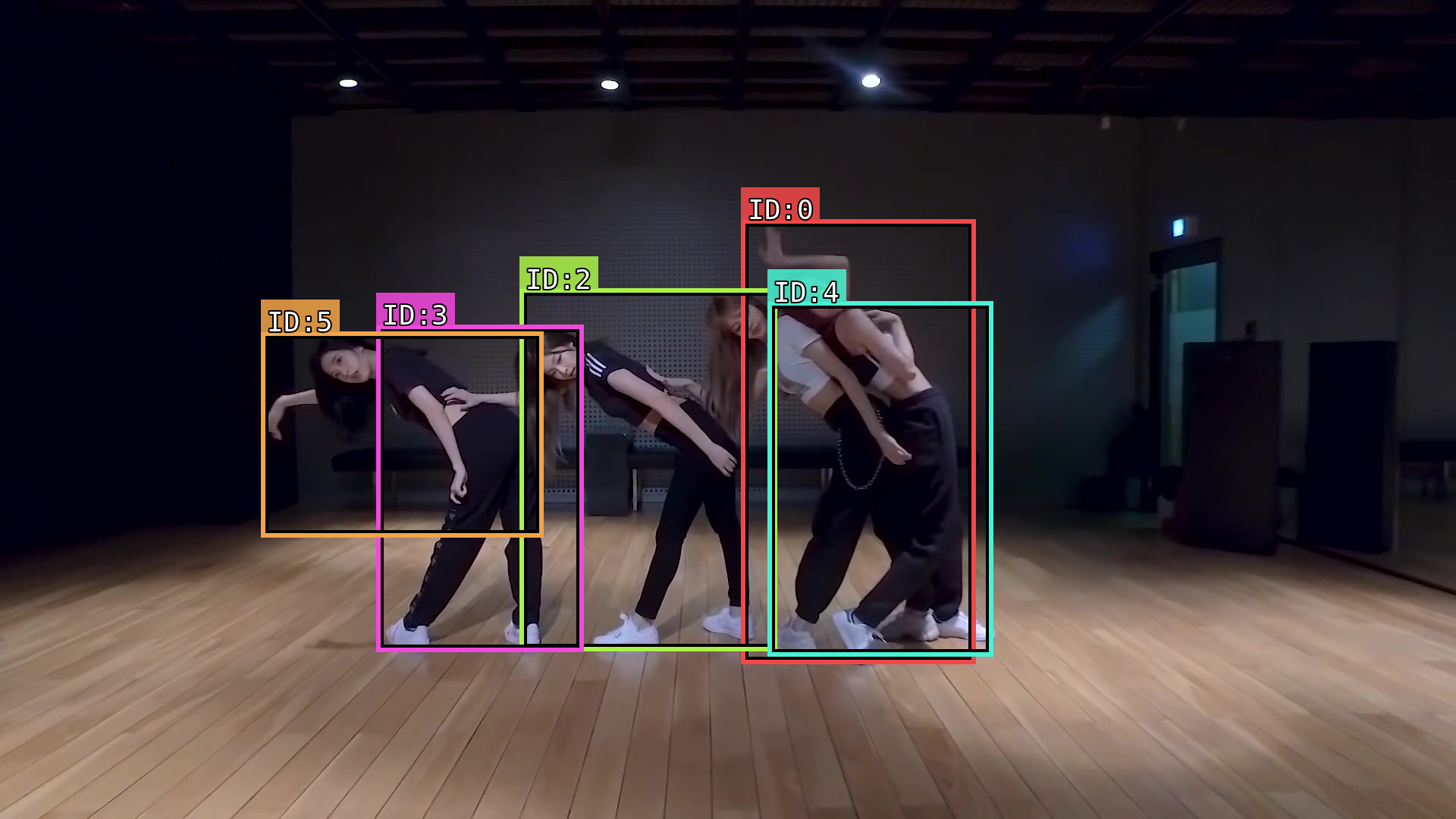}\hfill
        \includegraphics[width=\figwidth]{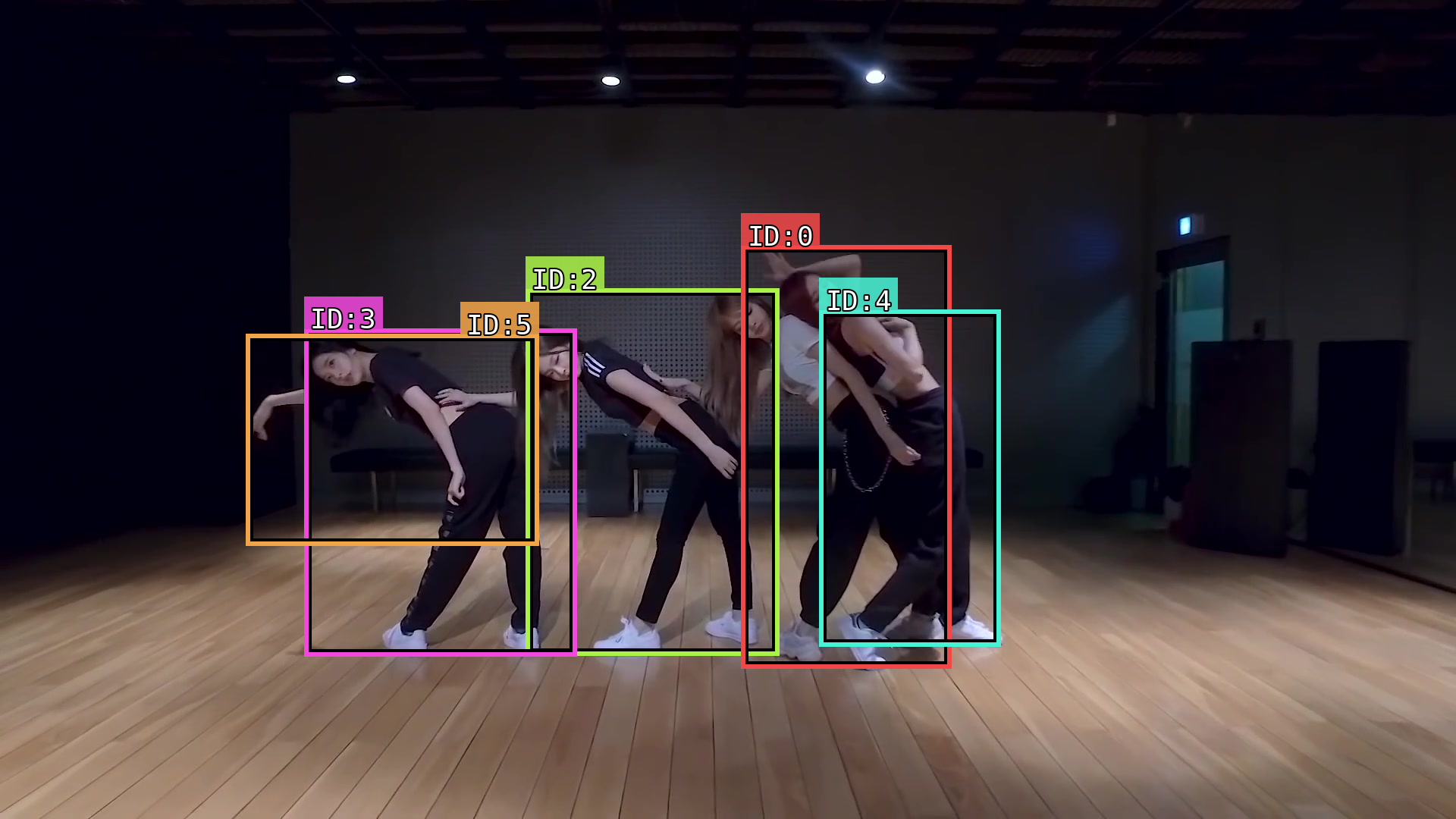}
        \subcaptext{(a) Random Sampling}
    \end{minipage}

    \vspace{2mm}

    \begin{minipage}[t]{\linewidth}
        \centering
        \includegraphics[width=\figwidth]{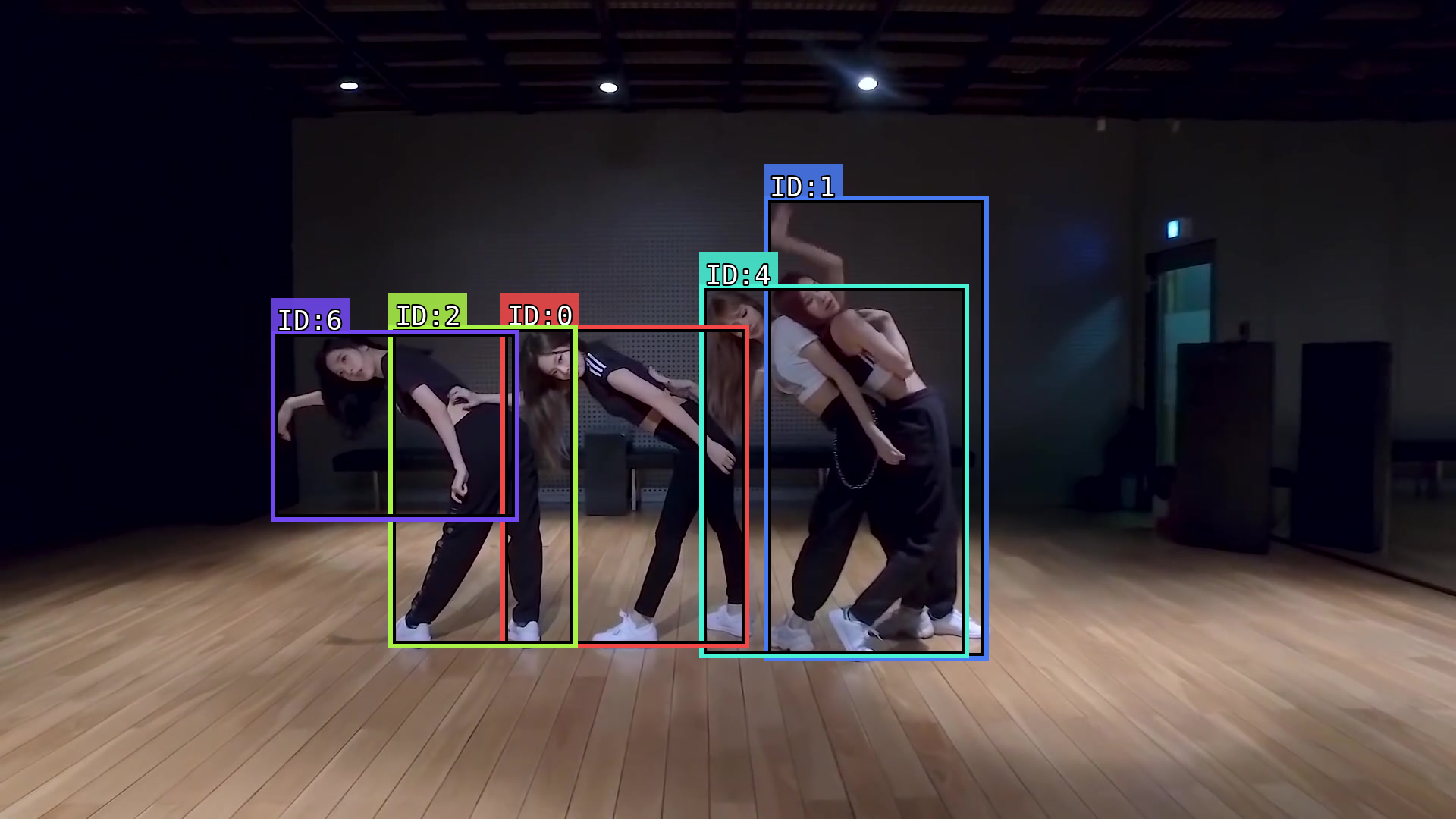}\hfill
        \includegraphics[width=\figwidth]{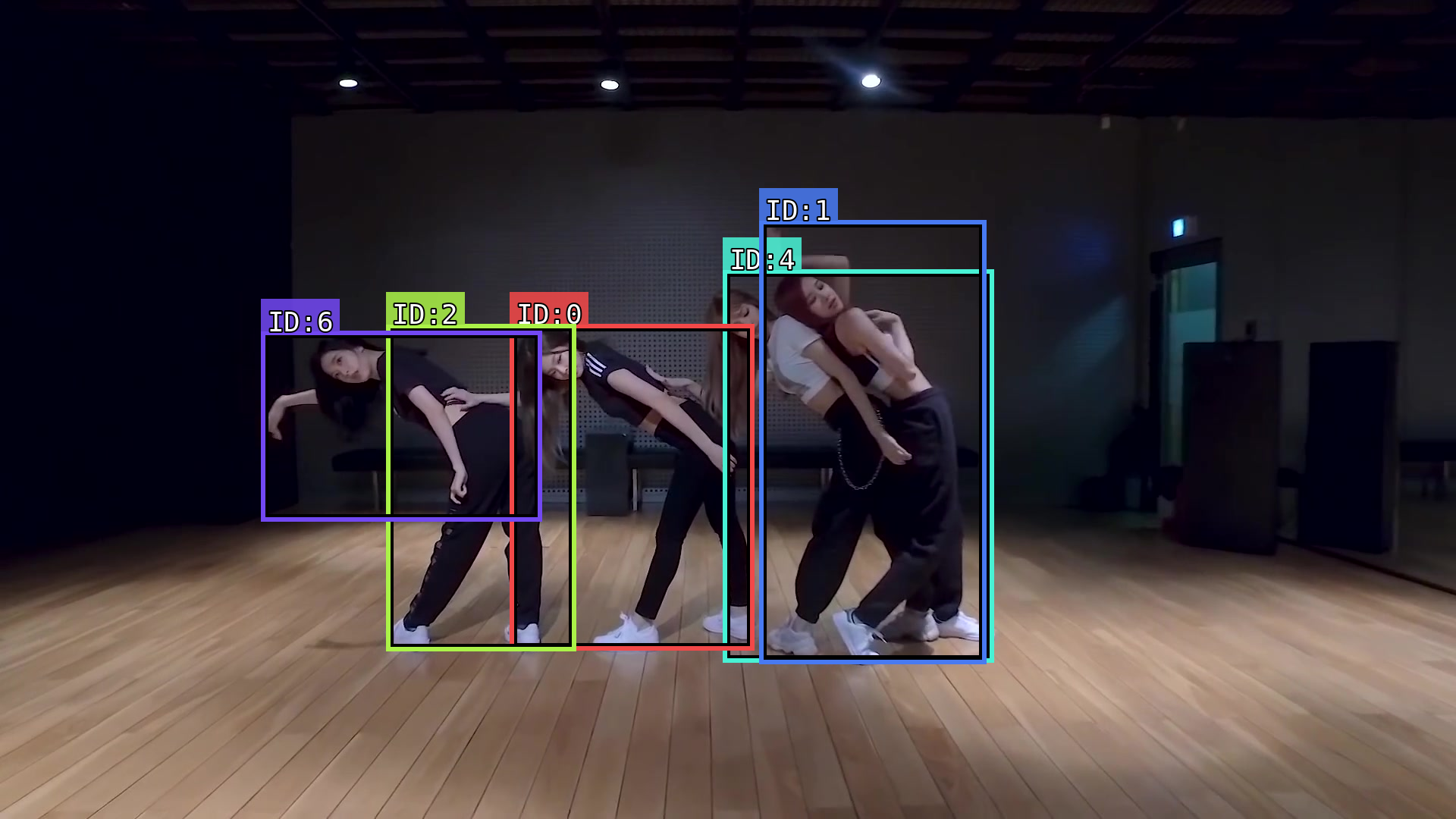}\hfill
        \includegraphics[width=\figwidth]{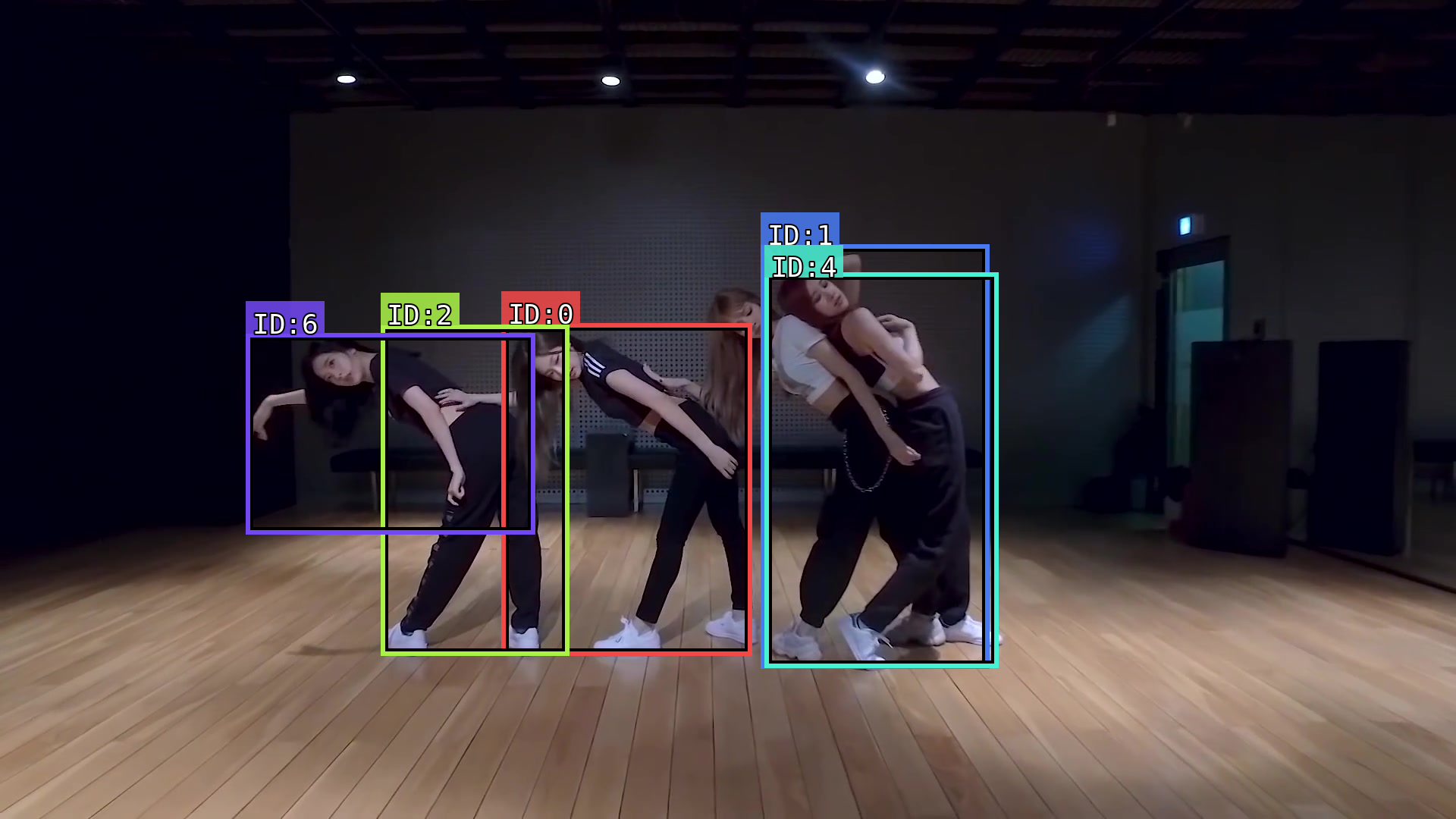}
        \subcaptext{(b) Full Supervision}
    \end{minipage}
    
    \vspace{2mm}

    \begin{minipage}[t]{\linewidth}
        \centering
        \includegraphics[width=\figwidth]{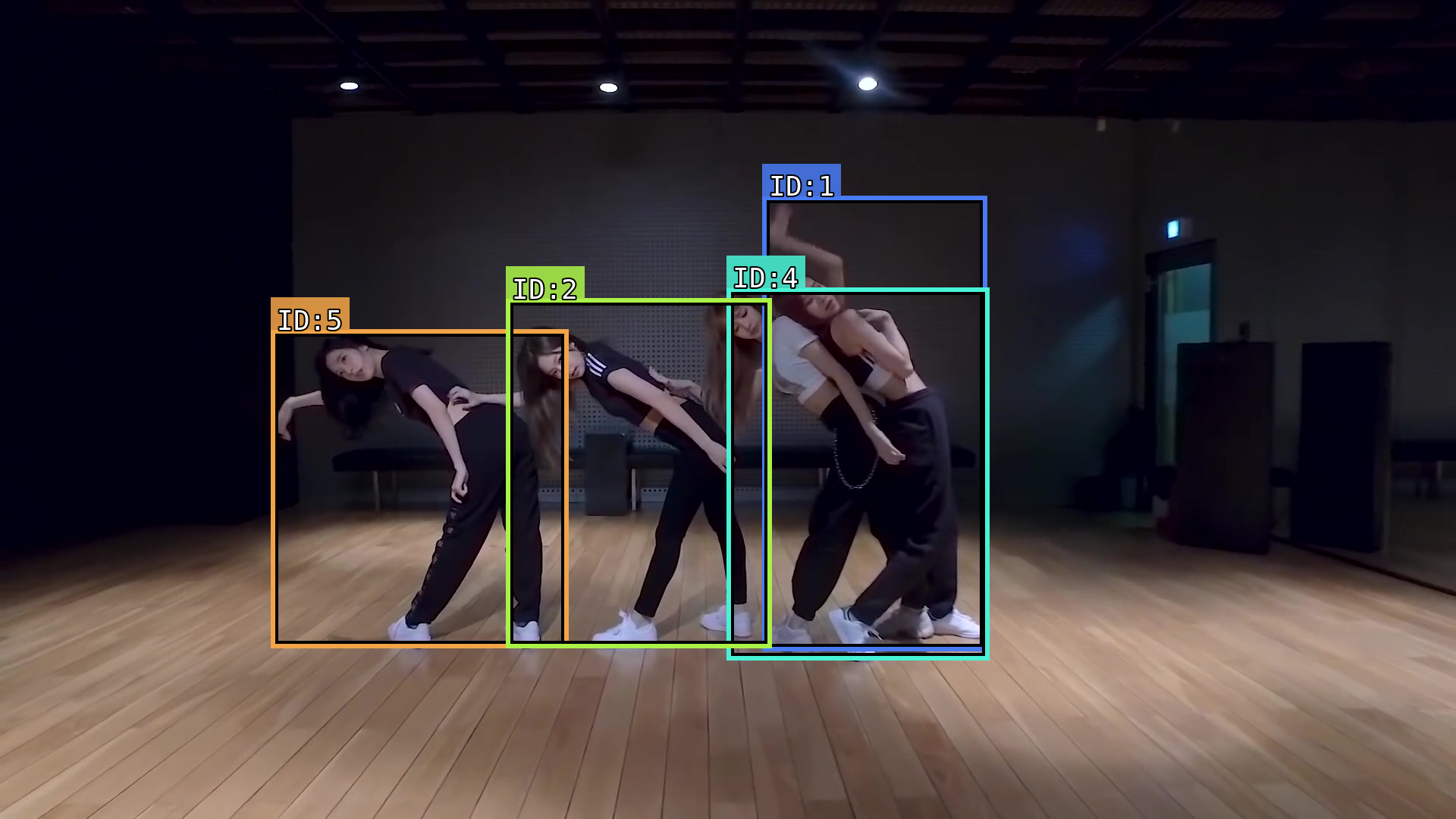}\hfill
        \includegraphics[width=\figwidth]{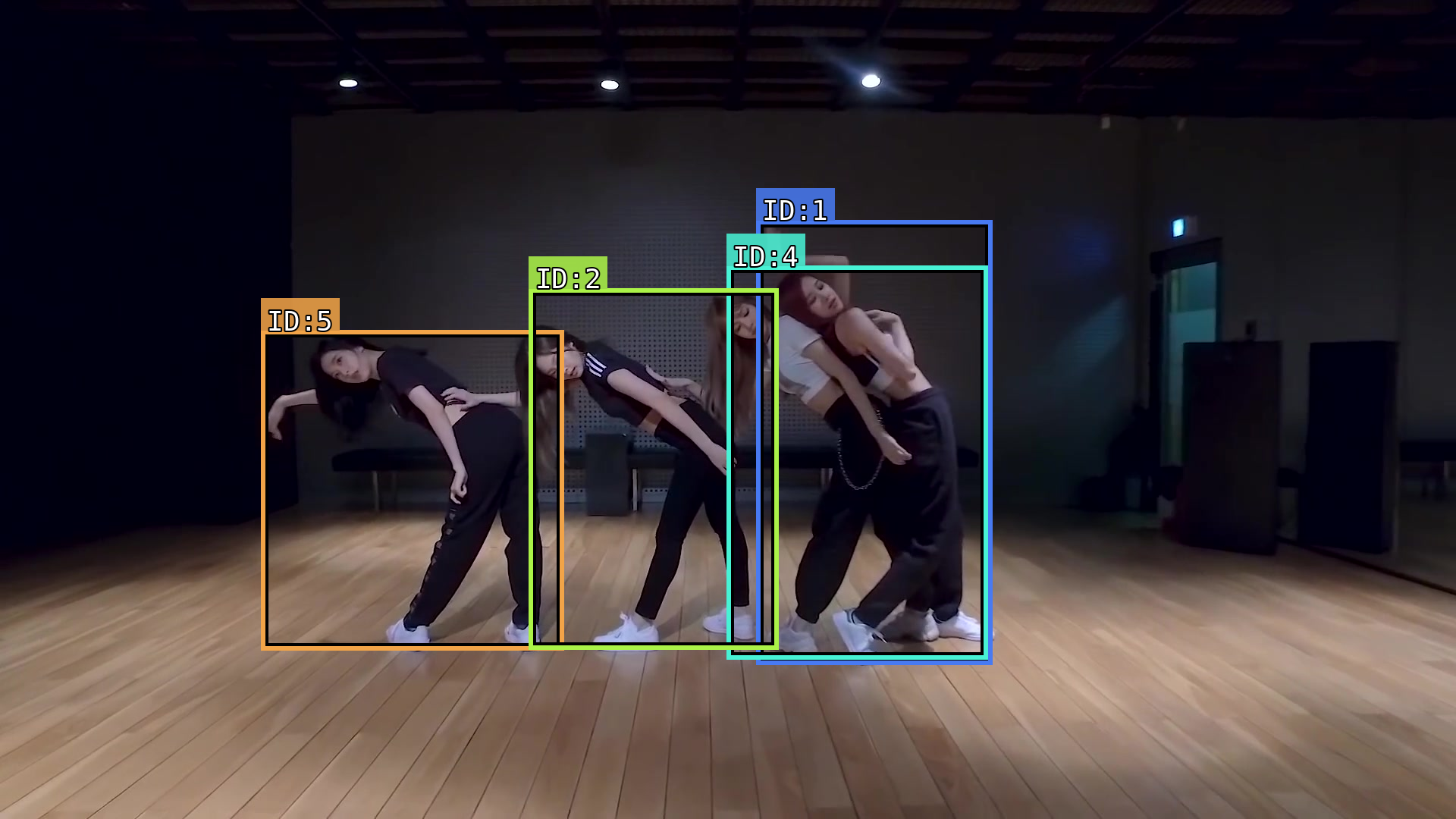}\hfill
        \includegraphics[width=\figwidth]{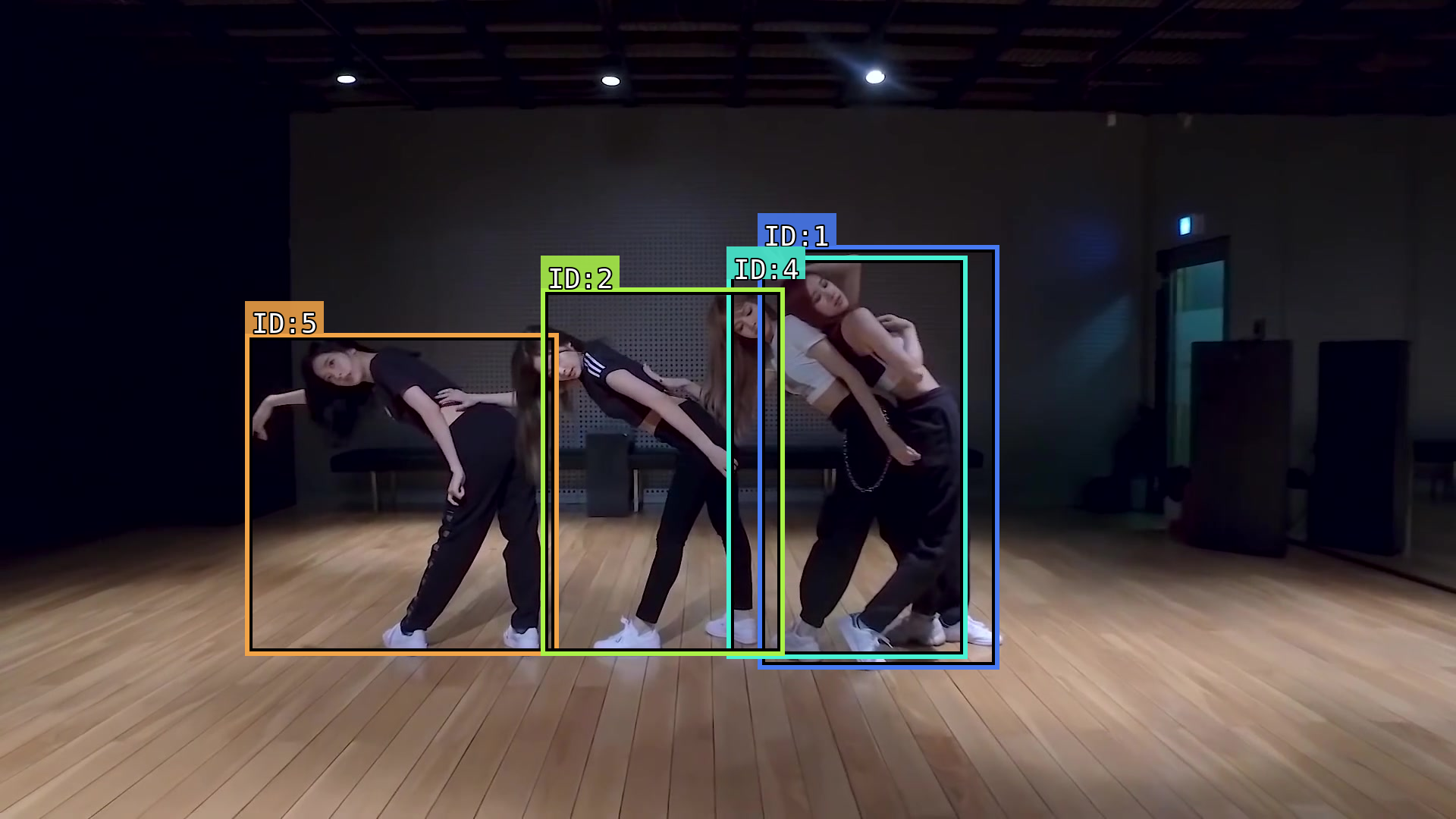}
        \subcaptext{(c) CUTAL (Ours)}
    \end{minipage}

    \vspace{3mm}

    \begin{minipage}[t]{\linewidth}
        \centering
        \includegraphics[width=\figwidth]{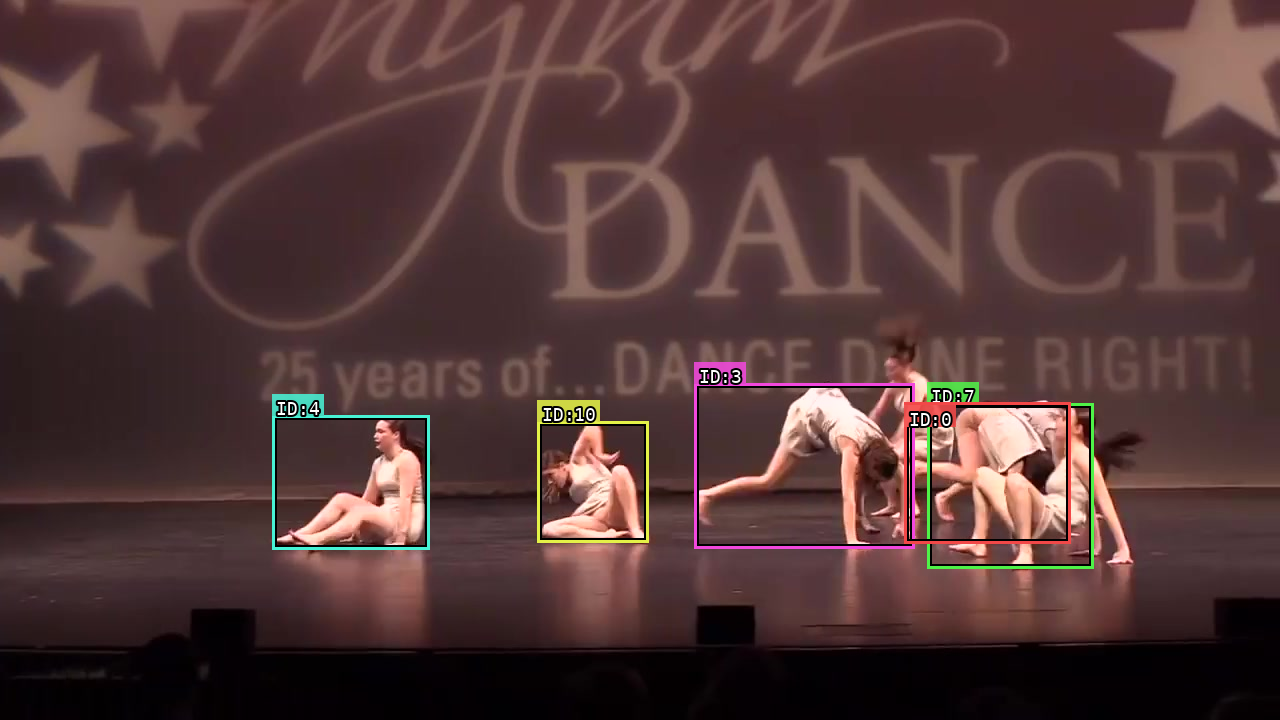}\hfill
        \includegraphics[width=\figwidth]{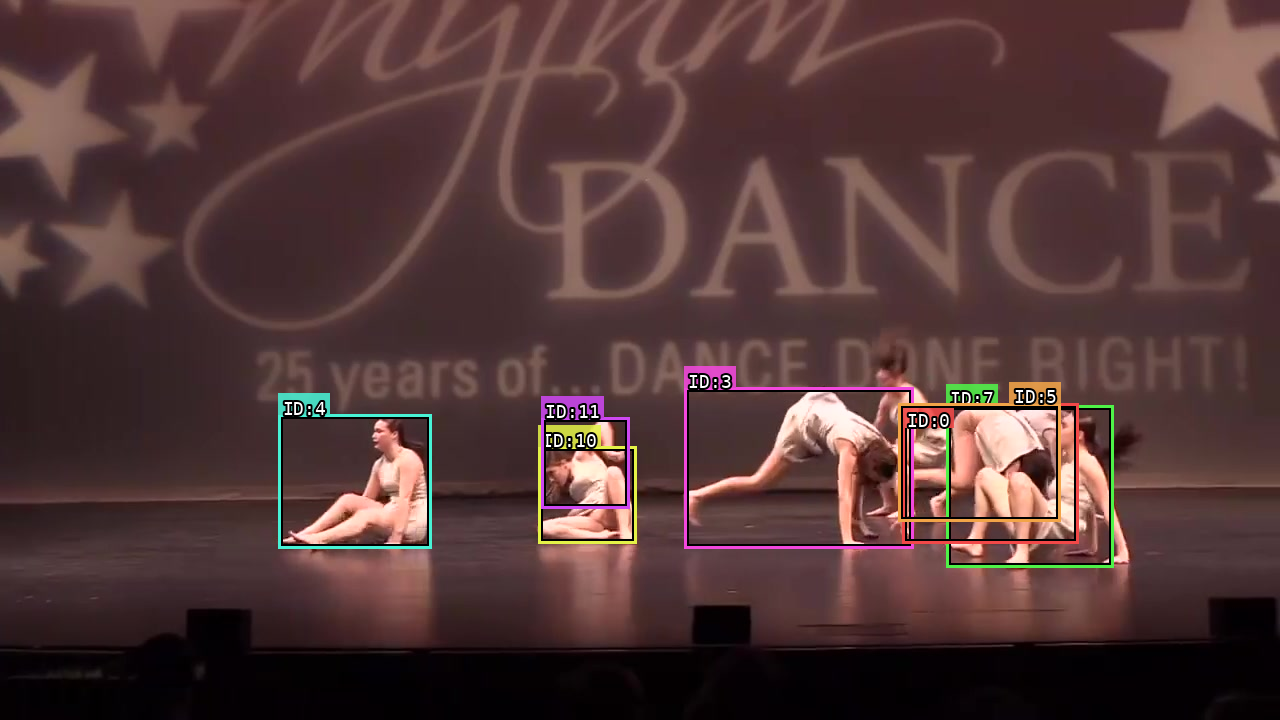}\hfill
        \includegraphics[width=\figwidth]{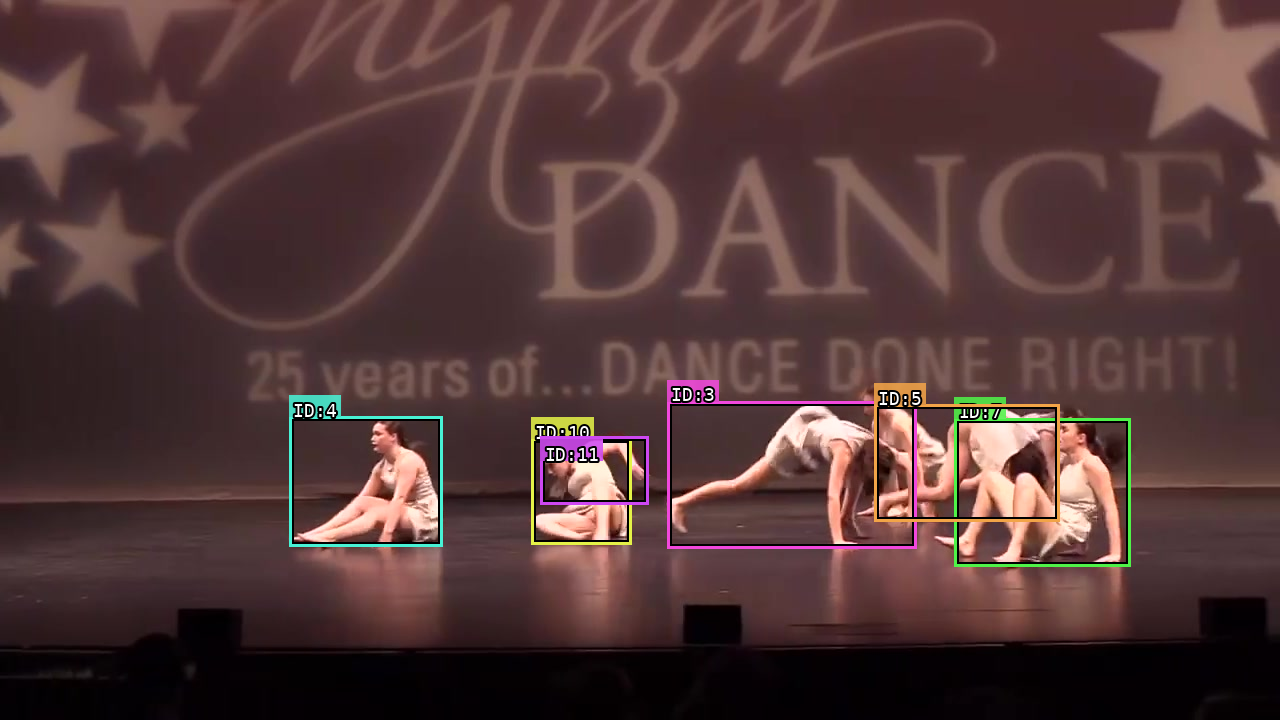}
        \subcaptext{(d) Random Sampling}
    \end{minipage}

    \vspace{2mm}

    \begin{minipage}[t]{\linewidth}
        \centering
        \includegraphics[width=\figwidth]{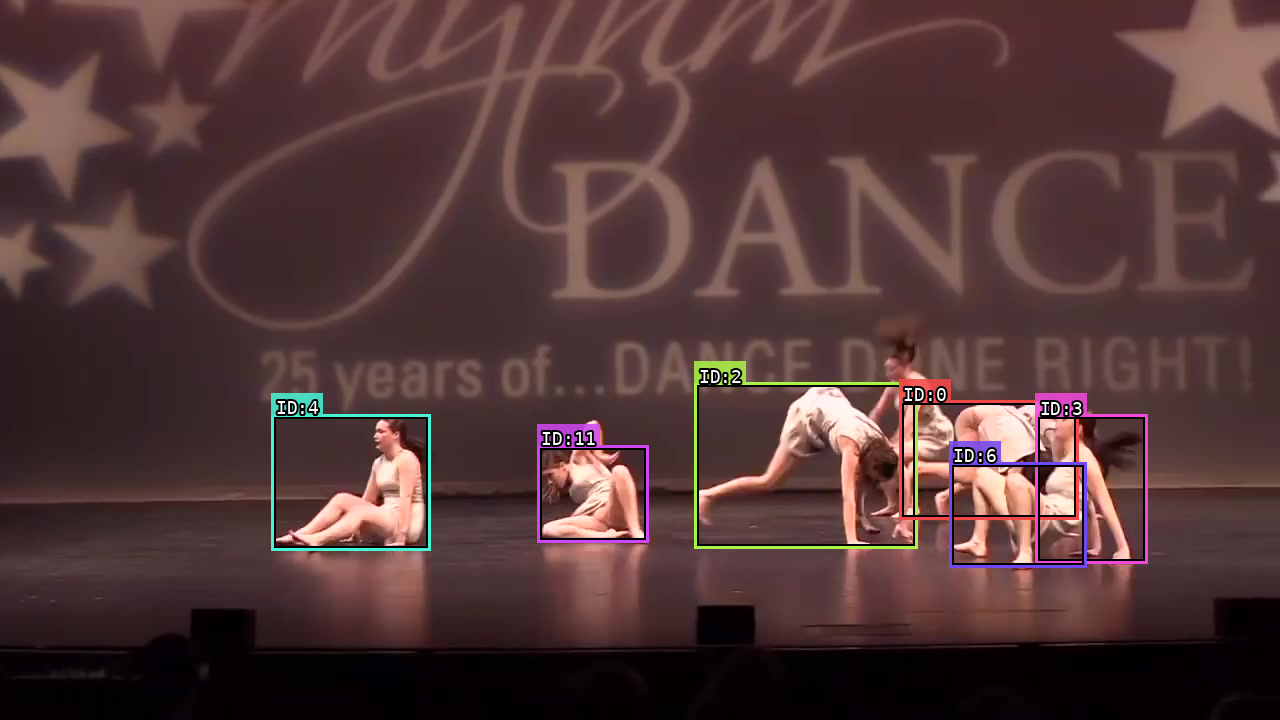}\hfill
        \includegraphics[width=\figwidth]{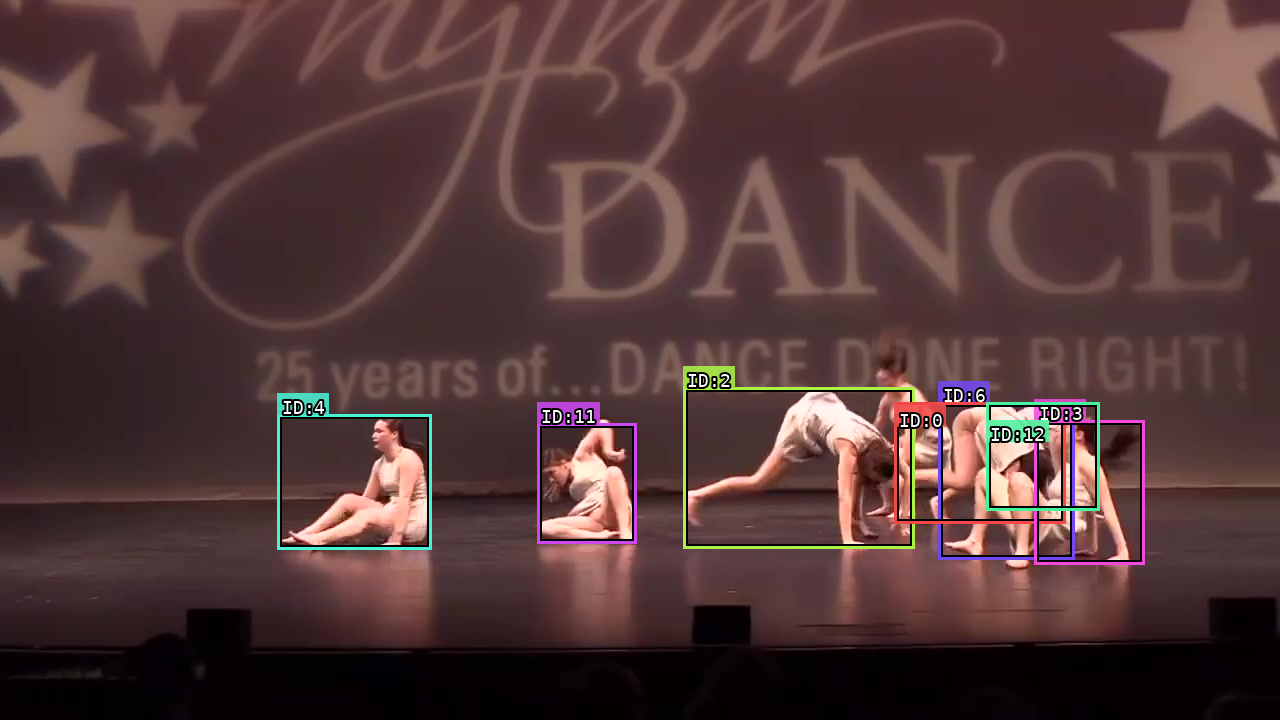}\hfill
        \includegraphics[width=\figwidth]{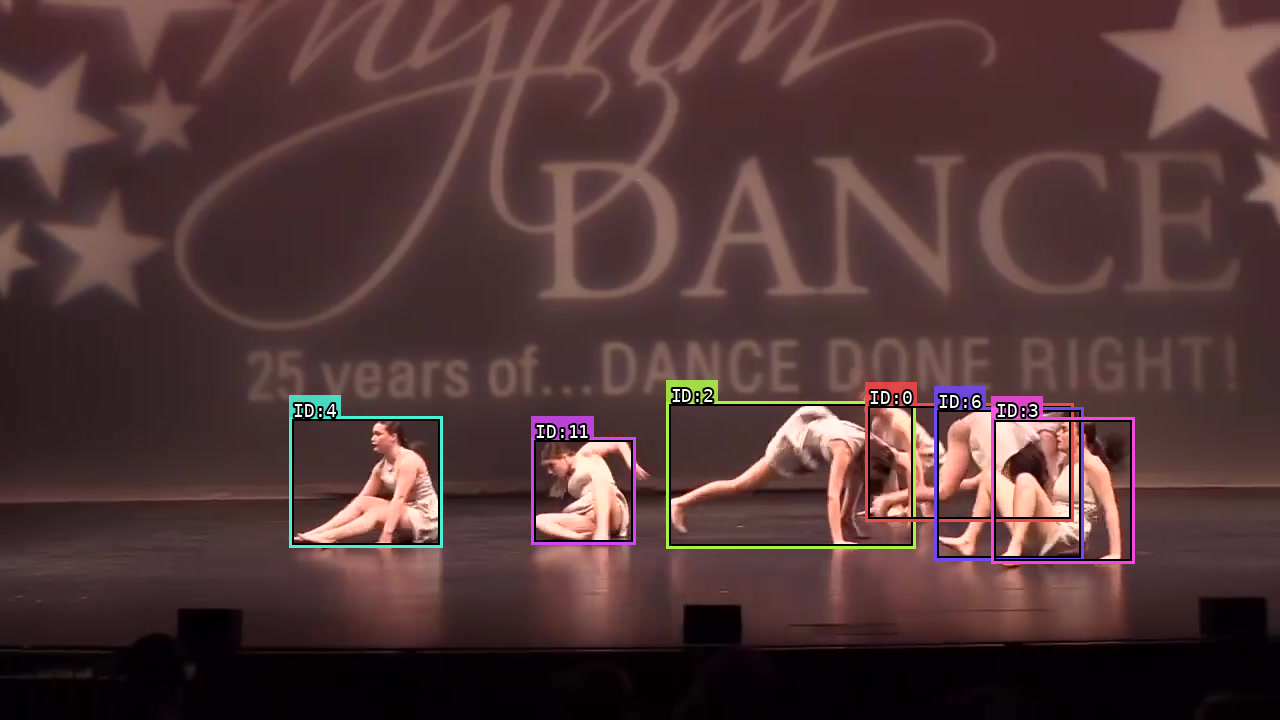}
        \subcaptext{(e) Full Supervision}
    \end{minipage}

    \vspace{2mm}

    \begin{minipage}[t]{\linewidth}
        \centering
        \includegraphics[width=\figwidth]{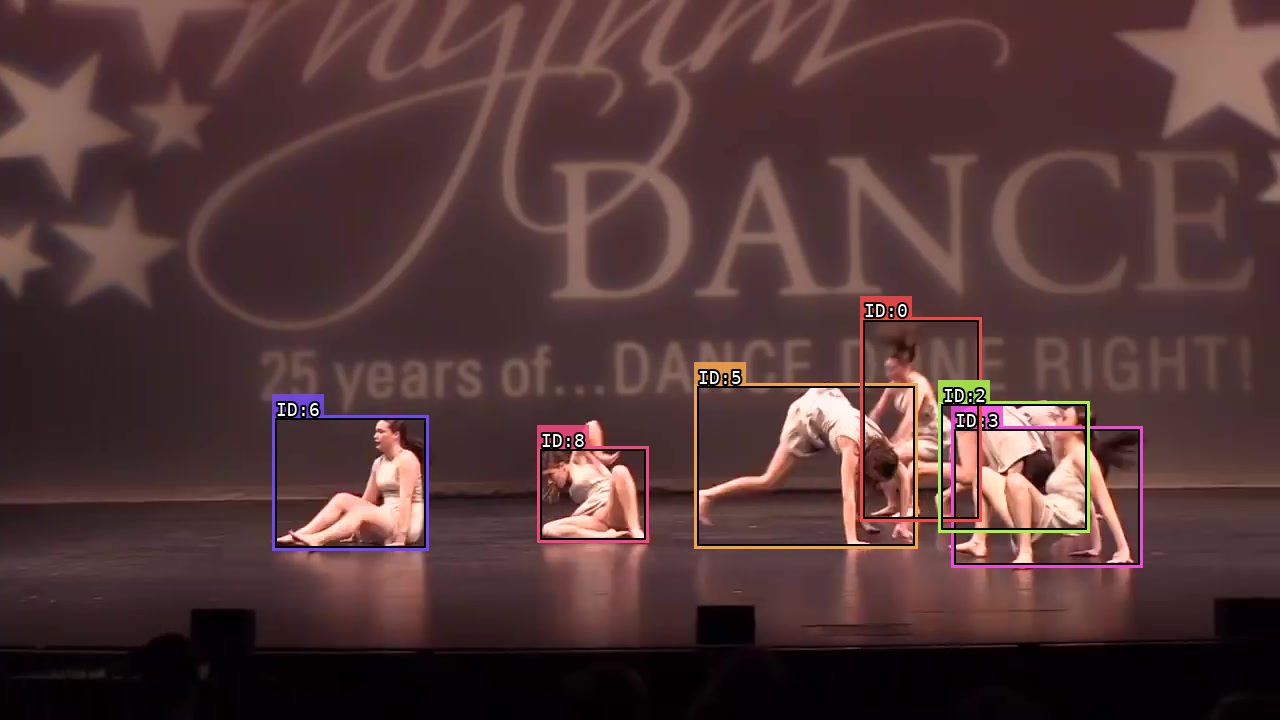}\hfill
        \includegraphics[width=\figwidth]{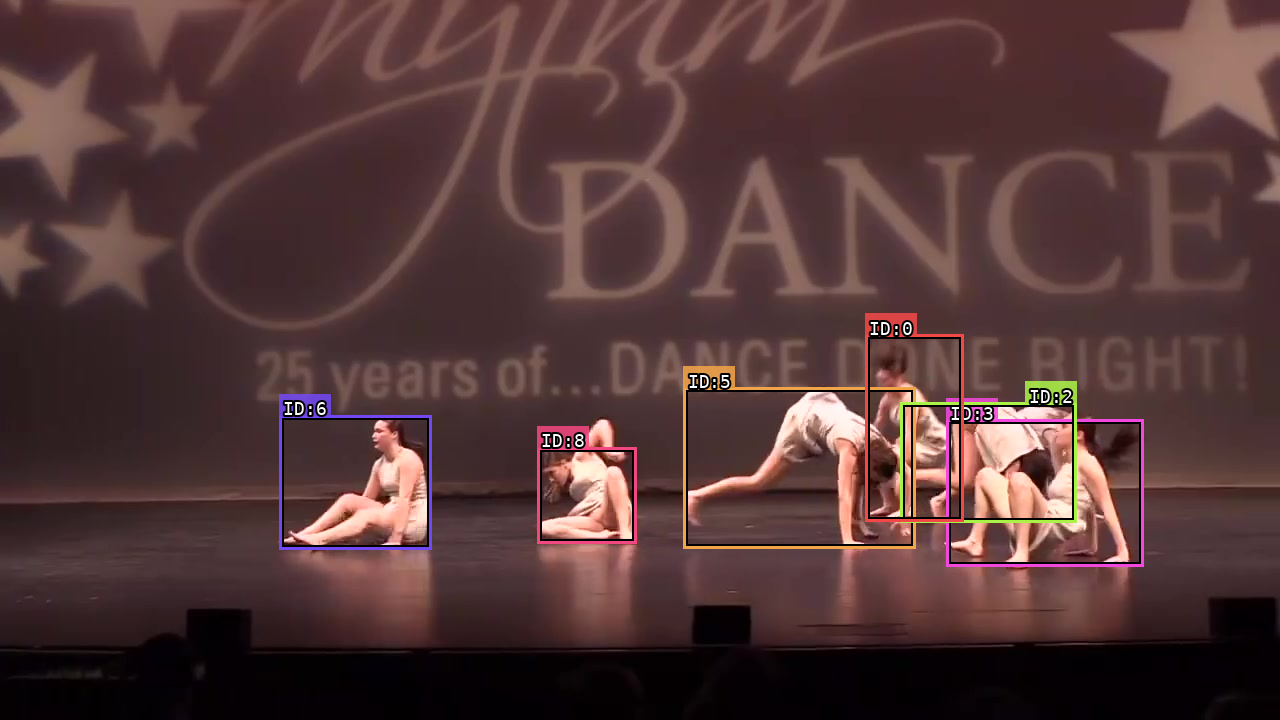}\hfill
        \includegraphics[width=\figwidth]{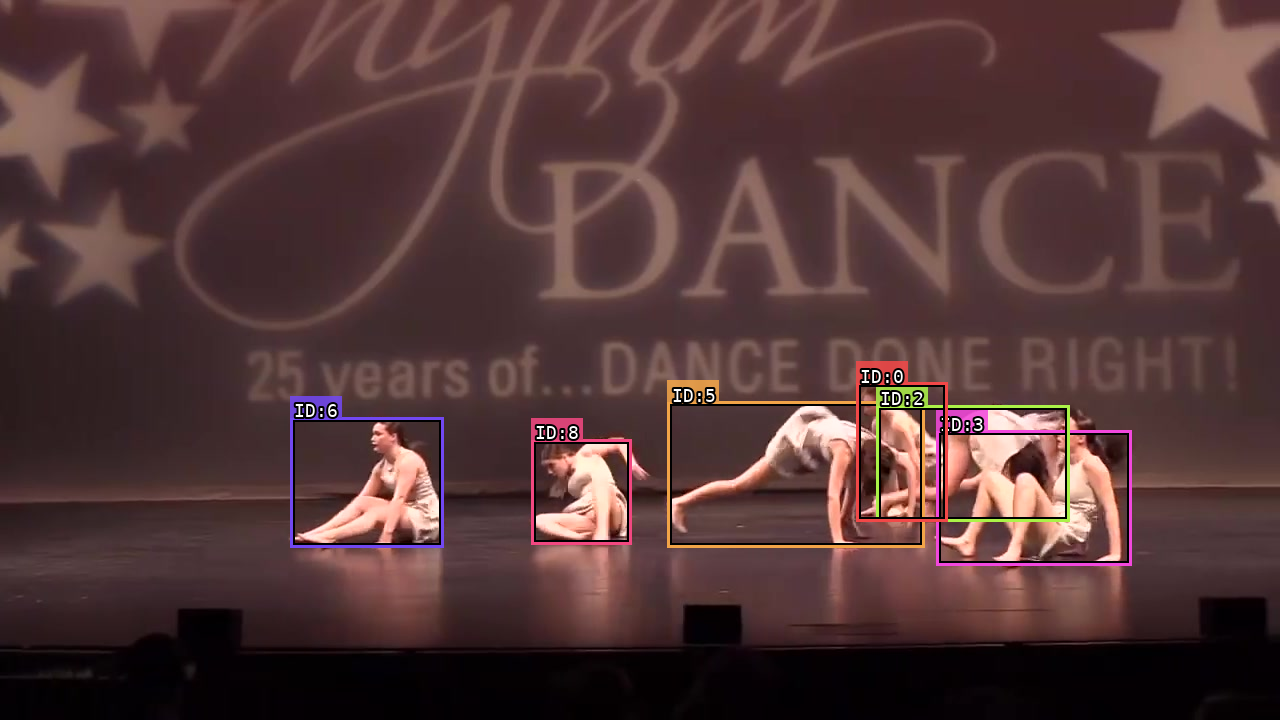}
        \subcaptext{(f) CUTAL (Ours)}
    \end{minipage}

    \caption{\textbf{Qualitative comparison on two sequences from the DanceTrack validation set.}
    Panels (a)--(c) show \texttt{dancetrack0005}, and panels (d)--(f) show \texttt{dancetrack0019}.}
    \label{fig:qual_dancetrack_all}
\end{figure*}

\noindent Figure~\ref{fig:qual_dancetrack_all} shows qualitative tracking examples of MeMOTR on two clips from the DanceTrack validation set.
Random sampling and CUTAL are trained with 50\% labeled data, while Full Supervision uses the full training set.
Random sampling often leads to missed detections and identity switches under close interactions and occlusions.
In contrast, CUTAL yields more stable tracks and identity assignments.

\section{Ablation Study}
\label{sec:supp_ablation}
\noindent Table~\ref{tab:cutal_ablation} reports the ablation results of MeMOTR on the DanceTrack validation set (5\%+5\% schedule).
Our full method performs best in Round 2 and Round 3, while remaining competitive in Round 1.
The smaller gain in Round 1 is likely due to the cold start effect \cite{chen2024making}, where training on only 5\% of the data yields unstable predictions and noisier uncertainty estimates.
We also test an additive score aggregation by replacing the multiplication in Eq.~(6) in the main paper with summation (\textit{Sum Agg.}).
Although \textit{Sum Agg.} is competitive in Round 1, multiplicative aggregation performs better in later rounds.

\begin{table}[t]
  \centering
  \caption{\textbf{Ablation study on uncertainty components, temporal diversity sampling, and score aggregation.}
  \textit{w/o Var.}, \textit{w/o Ent.}, and \textit{w/o BiDir.} remove ID-linked entropy variation, mean entropy, and bidirectional inconsistency, respectively.
  \textit{w/o Temp.} removes temporal diversity sampling.
  \textit{Sum Agg.} replaces the product aggregation of uncertainty scores with summation.}
  \label{tab:cutal_ablation}
  \vspace{1.3mm}
  \scriptsize
  \setlength{\tabcolsep}{3.2pt}
  \renewcommand{\arraystretch}{0.95}
  \begin{adjustbox}{width=0.8\linewidth,center}
  \begin{tabular}{clccc}
    \toprule
    Round & Method & HOTA & AssA & IDF1 \\
    \midrule
    \multirow{6}{*}{1}
    & CUTAL w/o Var.   & 53.18 & 40.24 & 54.82 \\
    & CUTAL w/o Ent.   & 53.57 & 41.42 & 56.38 \\
    & CUTAL w/o BiDir. & \textbf{54.46} & \textbf{42.34} & \textbf{57.47} \\
    & CUTAL w/o Temp.  & 53.76 & 41.67 & 56.29 \\
    & CUTAL (Sum Agg.) & 53.98 & 41.51 & \underline{56.61} \\
    & CUTAL (Ours)     & \underline{54.05} & \underline{41.72} & 56.33 \\
    \midrule
    \multirow{6}{*}{2}
    & CUTAL w/o Var.   & 54.94 & 42.54 & 57.36 \\
    & CUTAL w/o Ent.   & 55.43 & 42.81 & 56.95 \\
    & CUTAL w/o BiDir. & \underline{56.80} & 44.78 & \underline{59.15} \\
    & CUTAL w/o Temp.  & 56.72 & \underline{45.06} & 59.05 \\
    & CUTAL (Sum Agg.) & 55.46 & 43.25 & 57.54 \\
    & CUTAL (Ours)     & \textbf{56.91} & \textbf{45.17} & \textbf{59.24} \\
    \midrule
    \multirow{6}{*}{3}
    & CUTAL w/o Var.   & 57.30 & 45.47 & 59.24 \\
    & CUTAL w/o Ent.   & 57.46 & 45.39 & 59.94 \\
    & CUTAL w/o BiDir. & 58.21 & \underline{47.33} & \underline{60.94} \\
    & CUTAL w/o Temp.  & 57.72 & 46.55 & 60.35 \\
    & CUTAL (Sum Agg.) & \underline{58.23} & 46.83 & 60.64 \\
    & CUTAL (Ours)     & \textbf{59.14} & \textbf{48.45} & \textbf{61.69} \\
    \bottomrule
  \end{tabular}
  \end{adjustbox}
\end{table}

\section{Visualization of Sampled Clips}
\label{sec:supp_vis}

\begin{figure*}[t]
    \centering

    \begin{minipage}[t]{0.48\linewidth}
        \centering
        \includegraphics[width=\linewidth]{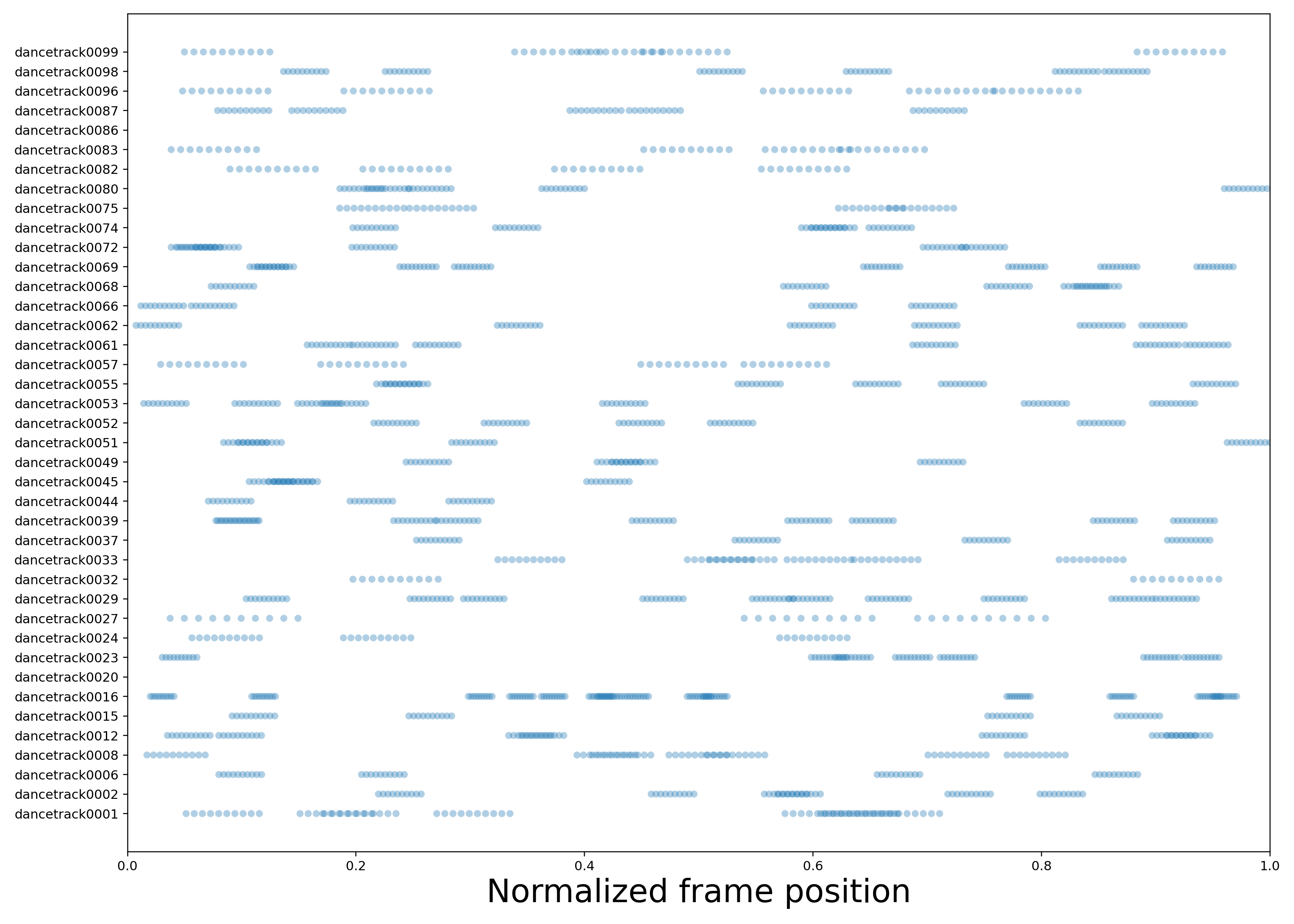}
        \subcaptext{(a) Random Sampling}
    \end{minipage}
    \hfill 
    \begin{minipage}[t]{0.48\linewidth}
        \centering
        \includegraphics[width=\linewidth]{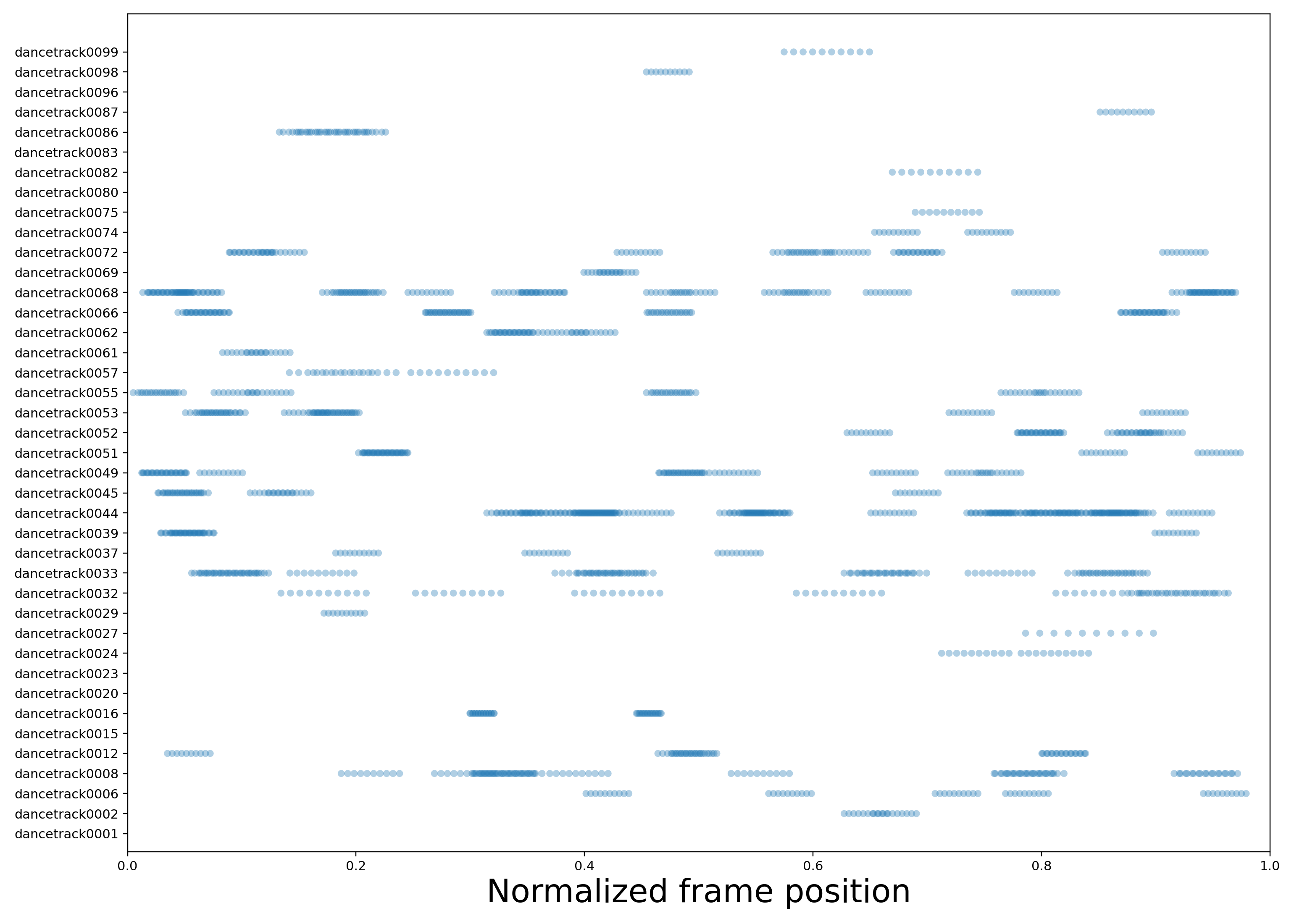}
        \subcaptext{(b) Entropy Sampling}
    \end{minipage}
    
    \vspace{2mm}

    \begin{minipage}[t]{0.48\linewidth}
        \centering
        \includegraphics[width=\linewidth]{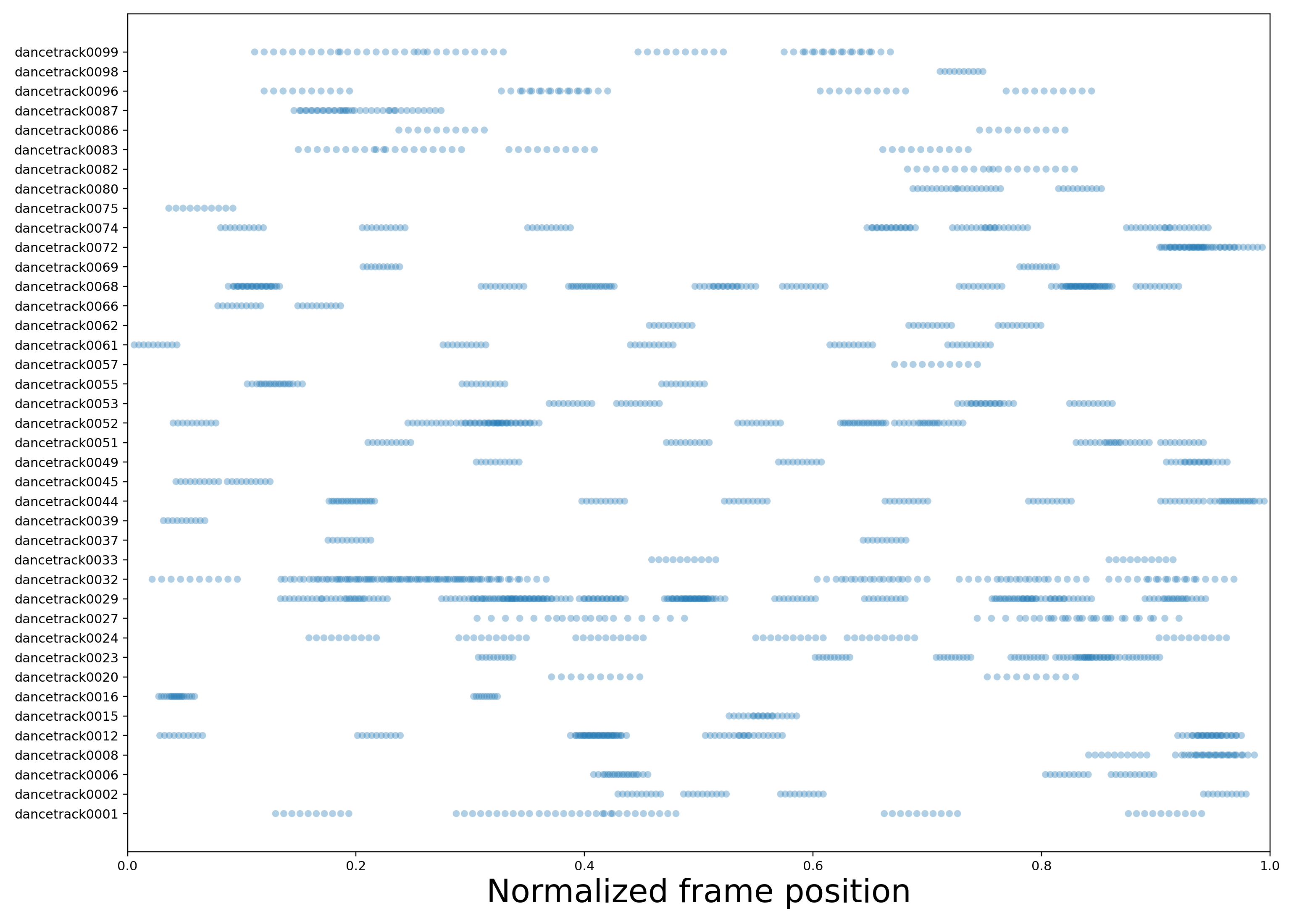}
        \subcaptext{(c) Core-set Sampling}
    \end{minipage}
    \hfill
    \begin{minipage}[t]{0.48\linewidth}
        \centering
        \includegraphics[width=\linewidth]{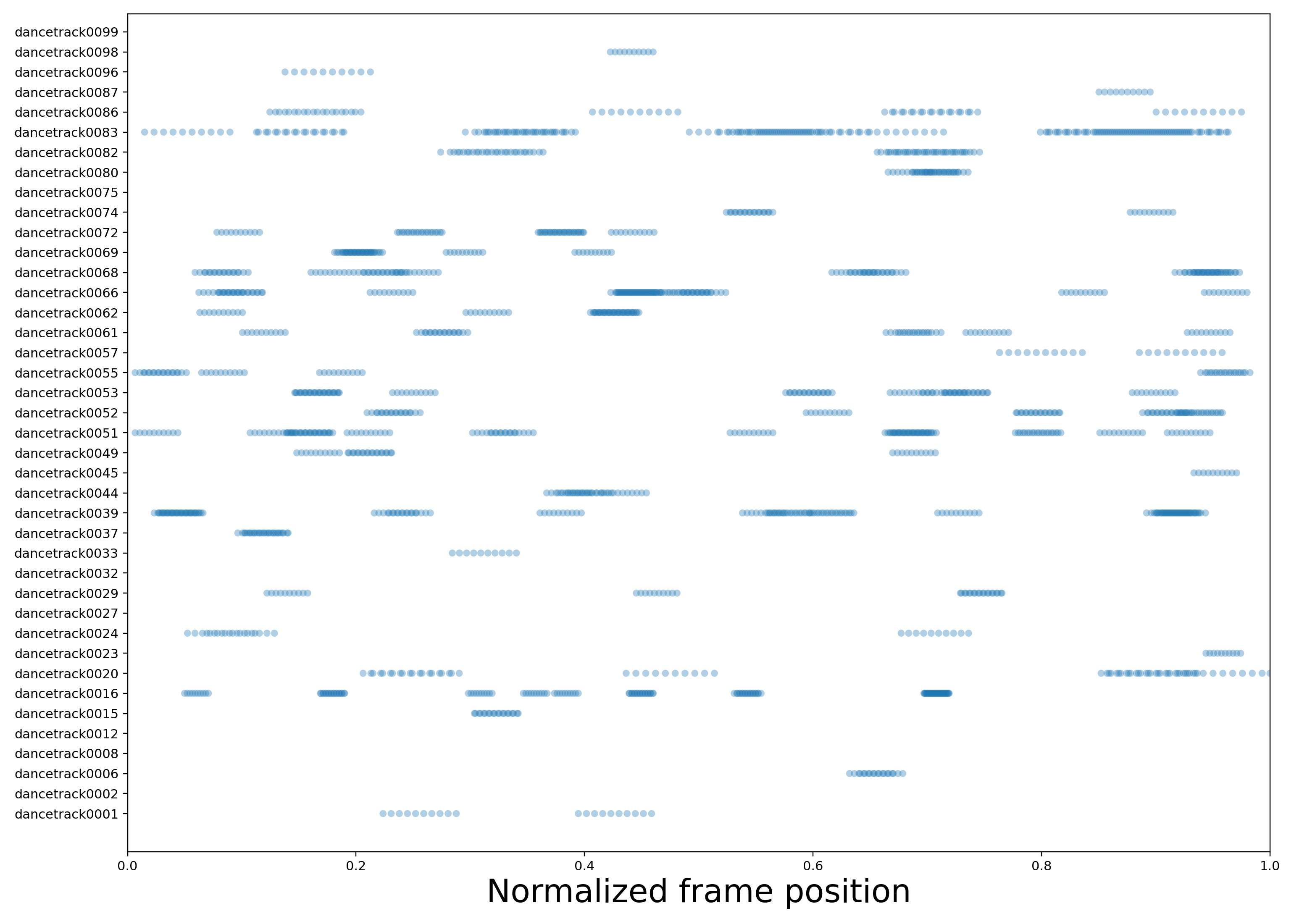}
        \subcaptext{(d) Ours w/o Temporal Diversity}
    \end{minipage}

    \vspace{2mm}

    \begin{minipage}[t]{0.48\linewidth}
        \centering
        \includegraphics[width=\linewidth]{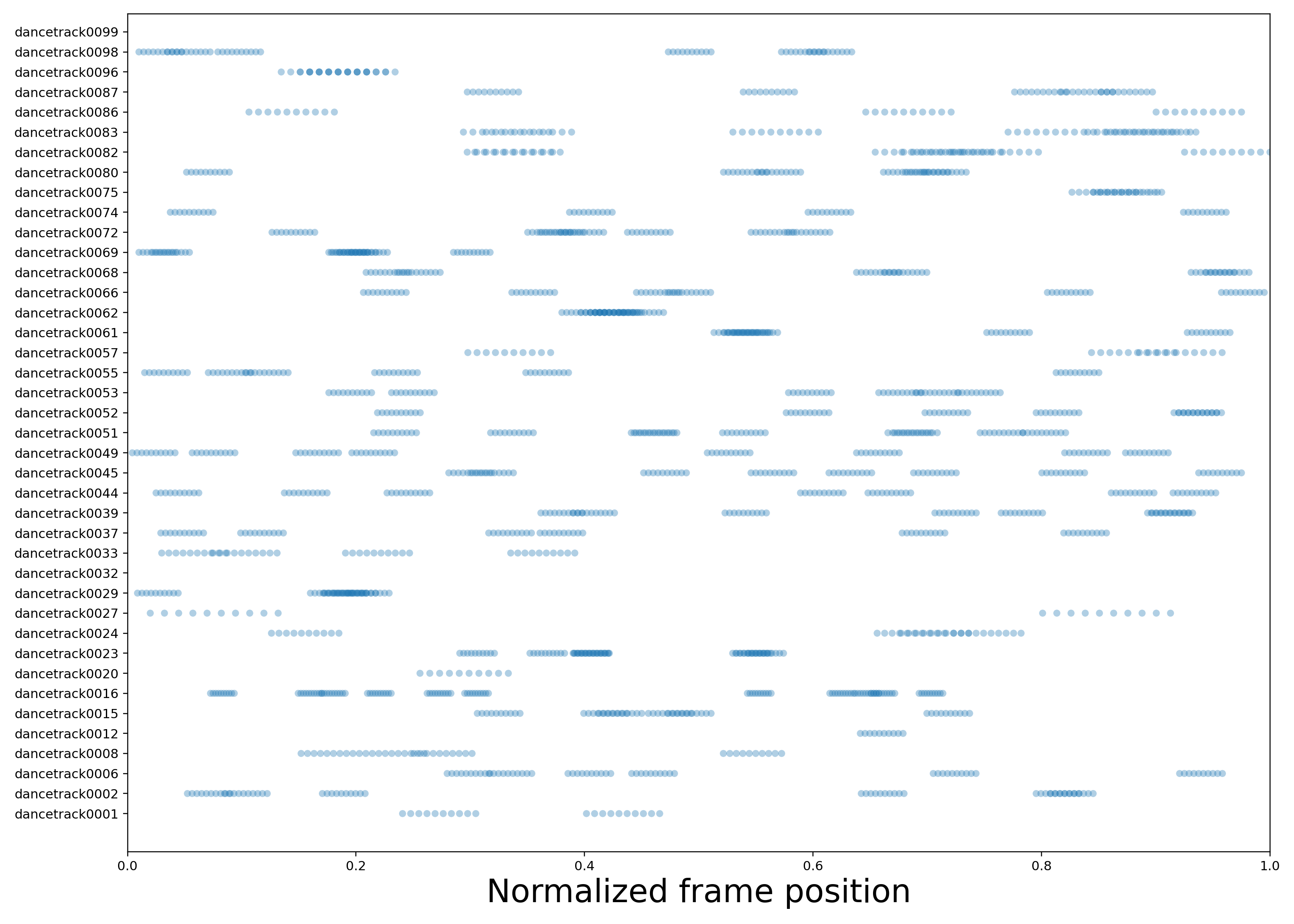}
        \subcaptext{(e) Ours (Full Method)}
    \end{minipage}

    \caption{\textbf{Visualization of sampled clips for training SambaMOTR on the DanceTrack dataset.}
    The x-axis represents the normalized frame position, and the y-axis corresponds to different video sequences in the dataset.
    Panels (a)--(c) show baseline methods, (d) shows our method without temporal diversity, and (e) shows our full proposed method.
    By comparing (d) and (e), we observe that the temporal diversity module effectively distributes samples across the timeline.}
    \label{fig:vis_sampling_comparison}
\end{figure*}

\noindent We visualize how different sampling strategies distribute selected clips along the timeline.
Figure~\ref{fig:vis_sampling_comparison} shows the normalized frame positions of sampled clips for training SambaMOTR on DanceTrack in Round 1.
Entropy sampling (b) yields selections concentrated in limited temporal regions.
In contrast, CUTAL (e) spreads selections more broadly than the variant without temporal diversity (d), reducing temporal redundancy while still selecting high-uncertainty candidates.

\section{Discussion}
\label{sec:discussion}

\noindent\textbf{Cold start in early rounds.}
In the first round, the tracker is trained on very limited data, leading to unstable representations and unreliable uncertainty estimates in this cold-start regime~\cite{chen2024making}.
Accordingly, performance may fluctuate and simple heuristics such as Entropy can occasionally outperform CUTAL, e.g., MeMOTR on DanceTrack under the 20\%+10\% schedule in Table~\ref{tab:dancetrack_20p10_memotr_sambamotr_vertical}.
As more labels are acquired, uncertainty estimates become more reliable, and CUTAL provides more consistent gains by combining uncertainty with temporal diversification.

\noindent\textbf{Performance Gap in SambaMOTR.}
CUTAL approaches full-supervision performance with MeMOTR at 50\% labeled data, but a larger gap remains for SambaMOTR.
We attribute this mainly to the longer clip length under a fixed frame budget.
Since we annotate at most $b=\lfloor B/T\rfloor$ clips per round, SambaMOTR with $T{=}10$ acquires substantially fewer unique clips than MeMOTR with $T{=}4$, reducing coverage of distinct temporal contexts and interaction patterns.
Moreover, longer clips contain more internally correlated frames, so a larger fraction of the budget is spent on redundant observations with lower marginal information gain.

\noindent\textbf{Performance Analysis on SportsMOT.}
The performance gap between CUTAL and the baselines is smaller on SportsMOT than on DanceTrack.
We attribute this to differences in both task characteristics and dataset scale.
First, SportsMOT features fast object motion and camera movement, whereas DanceTrack is designed to stress identity association under frequent occlusions and visually similar targets, where selecting hard-to-associate clips can be particularly beneficial.
Second, under our training splits, SportsMOT has fewer training frames (approximately 28.5k) than DanceTrack (approximately 41.8k), which leaves less room for improvement from selective acquisition.
Consistent with this, even under the 5\%+5\% schedule, the baselines already achieve strong performance and the gains in later rounds become smaller.
{
    \small
    \putbib[main]
}
\end{bibunit}


\end{document}